\documentclass[letterpaper, 10 pt, conference]{ieeeconf}  % Comment this line out if you need a4paper

\IEEEoverridecommandlockouts                              % This command is only needed if 
\usepackage{amsmath} % assumes amsmath package installed
\usepackage{graphicx}
\usepackage{pgfplots}
\usepackage{tikz}
\usepgfplotslibrary{external} 
\usetikzlibrary{external,arrows.meta,patterns,ipe,positioning} 
\usepackage[font=footnotesize]{caption}
\usepackage[font=footnotesize]{subcaption}
\usepackage{xspace}

\title{\LARGE \bf
Learning the sense of touch in simulation: \\ a sim-to-real strategy for vision-based tactile sensing
}

\author{Carmelo Sferrazza, Thomas Bi and Raffaello D'Andrea
\thanks{The authors are members of the Institute for Dynamic Systems and Control, ETH Zurich, Switerland. Email correspondence to Carmelo Sferrazza
        {\tt\small csferrazza@ethz.ch}}%
}

\begin{document}
	
\maketitle

%\twocolumn[{%
%	\renewcommand\twocolumn[1][]{#1}%
%	\maketitle
%	\begin{center}
%		\captionsetup{type=figure}
%		\begin{tabular}[t]{ccc}
%			\subcaptionbox{The first}{\includegraphics[width=0.3\textwidth,height=0.3\textwidth]{example-image}}
%			& \subcaptionbox{The second}{\includegraphics[width=0.3\textwidth,height=0.3\textwidth]{example-image}}& \subcaptionbox{The first}{\includegraphics[width=0.3\textwidth,height=0.3\textwidth]{example-image}}\\[1cm]
%			\subcaptionbox{The third}{\includegraphics[width=0.3\textwidth,height=0.3\textwidth]{example-image}}
%			& \subcaptionbox{The last}{\includegraphics[width=0.3\textwidth,height=0.3\textwidth]{example-image}} & \subcaptionbox{The first}{\includegraphics[width=0.3\textwidth,height=0.3\textwidth]{example-image}} \\
%		\end{tabular}
%%		\subcaptionbox{Dataset generation as in \cite{sferrazza_fem}}{
%%			\scalebox{0.25}{\input{img/FEM_groundtruth.tex}}
%%		} \\
%%		\vskip\baselineskip
%%		\subcaptionbox{Dataset generation proposed here}{
%%			\scalebox{0.25}{\input{img/synthetic_training.tex}}
%%		} \hfill
%		\caption{A sequence of snapshots of the different steps of the pipeline described in this paper.}
%%		\label{fig:flow_diagram}
%%		\centering
%%		\includegraphics[width=\textwidth,height=5cm]{example-image}
%%		\captionof{figure}{Test caption}
%	\end{center}%
%}]

\thispagestyle{empty}
\pagestyle{empty}
%\begin{figure}[h]
%	\centering
%	\includegraphics[width=2\linewidth]{img/dummy.png}
%	\caption{Frames definition.}
%	\label{fig:frames_definition}
%\end{figure}

%%%%%%%%%%%%%%%%%%%%%%%%%%%%%%%%%%%%%%%%%%%%%%%%%%%%%%%%%%%%%%%%%%%%%%%%%%%%%%%%
\begin{abstract}
Data-driven approaches to tactile sensing aim to overcome the complexity of accurately modeling contact with soft materials. However, their widespread adoption is impaired by concerns about data efficiency and the capability to generalize when applied to various tasks. This paper focuses on both these aspects with regard to a vision-based tactile sensor, which aims to reconstruct the distribution of the three-dimensional contact forces applied on its soft surface. Accurate models for the soft materials and the camera projection, derived via state-of-the-art techniques in the respective domains, are employed to generate a dataset in simulation. A strategy is proposed to train a tailored deep neural network entirely from the simulation data. The resulting learning architecture is directly transferable across multiple tactile sensors without further training and yields accurate predictions on real data, while showing promising generalization capabilities to unseen contact conditions.
\end{abstract}

%%%%%%%%%%%%%%%%%%%%%%%%%%%%%%%%%%%%%%%%%%%%%%%%%%%%%%%%%%%%%%%%%%%%%%%%%%%%%%%%
\section{INTRODUCTION} \label{sec:introduction} 
\begin{figure}[h!]
%	\centering
	\subcaptionbox{Simulated indentation}{
		\setlength{\fboxsep}{0pt}
		\fbox{\includegraphics[height=0.25\columnwidth]{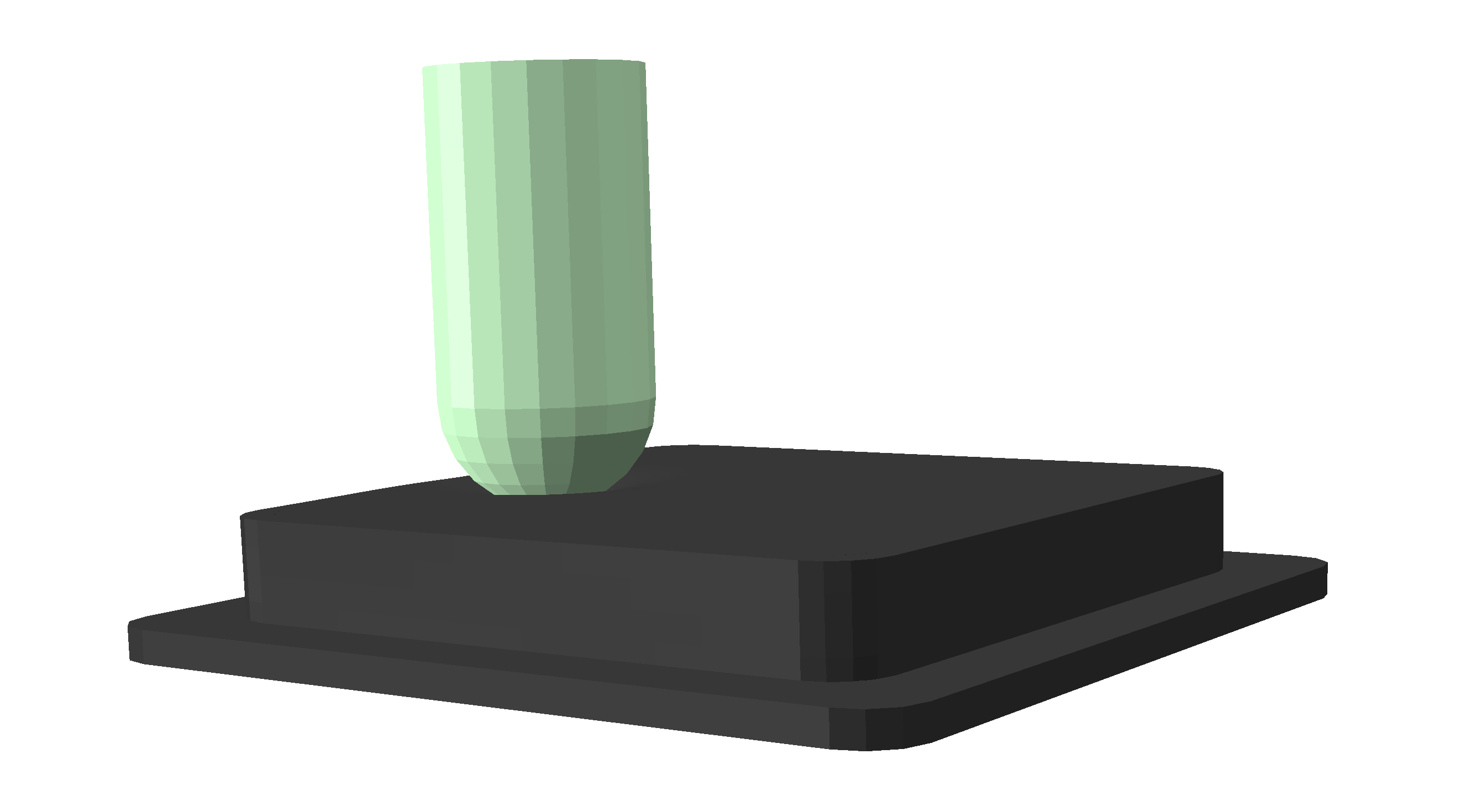}}
	} \hfill
	\subcaptionbox{Real-world indentation}{
		\setlength{\fboxsep}{0pt}
		\fbox{\includegraphics[height=0.25\columnwidth]{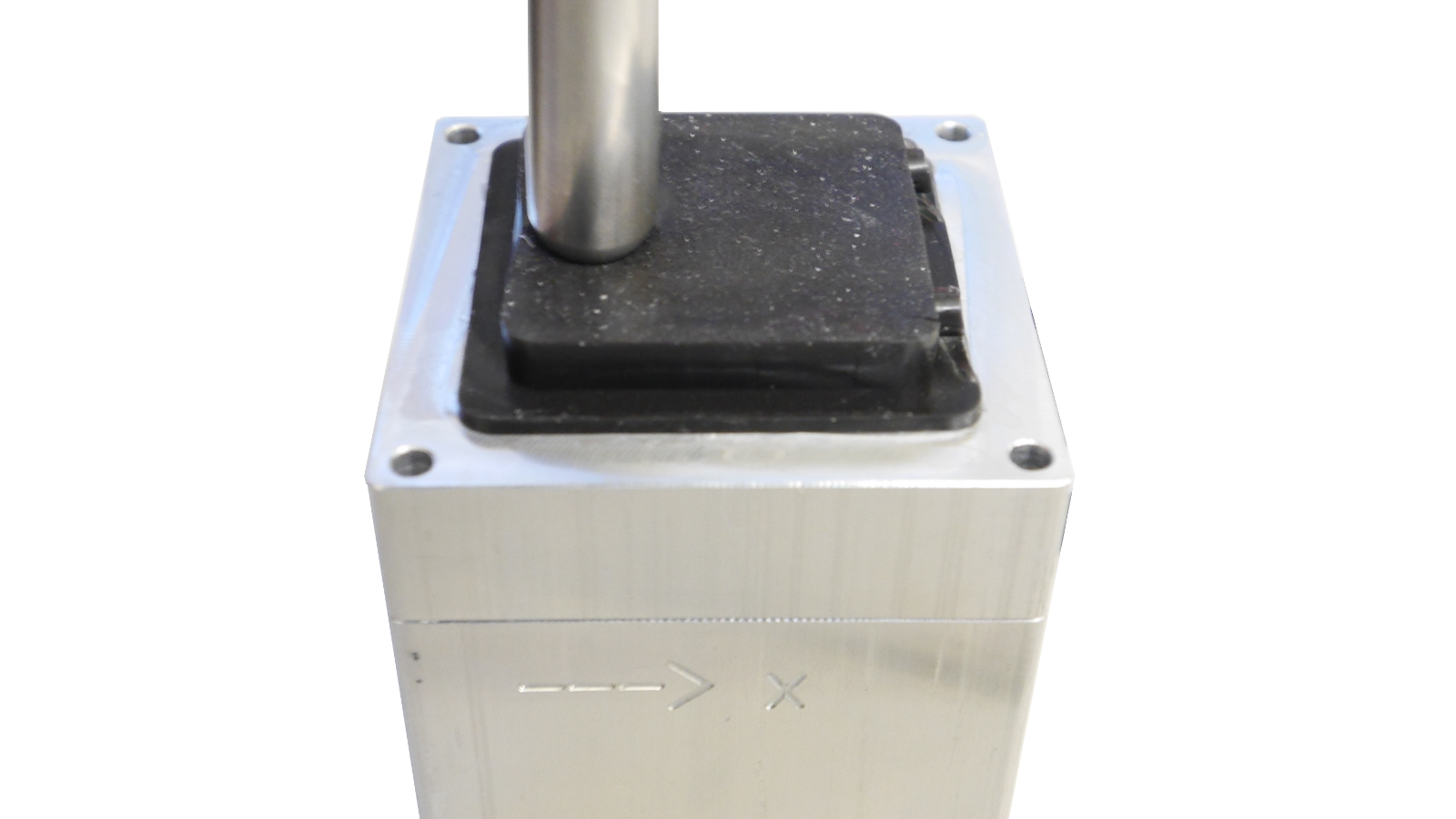}}
	}
 \vskip \baselineskip	\hspace{0.5mm}
	\subcaptionbox{Synthetic OF features}{
	\includegraphics[width=0.44\columnwidth]{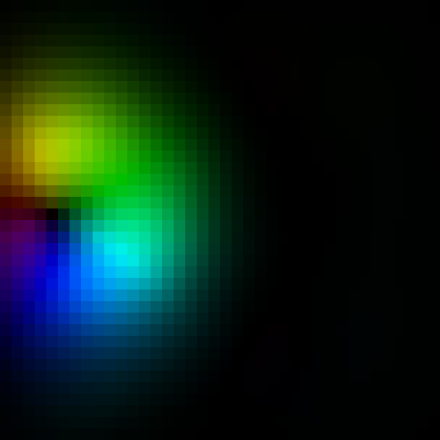}
} \hfill
\subcaptionbox{Real-world OF features}{
	\includegraphics[width=0.44\columnwidth]{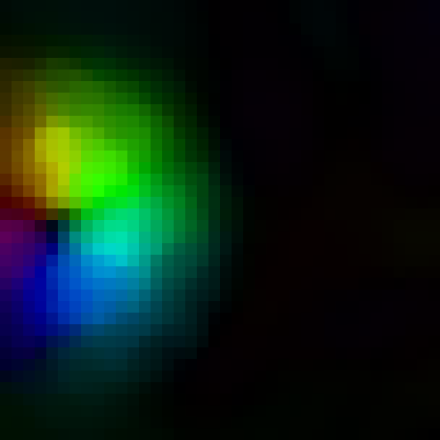}
}
 \vskip \baselineskip	
\hspace{-0.3cm}
	\subcaptionbox{Ground truth}{
	\includegraphics[height=0.45\columnwidth]{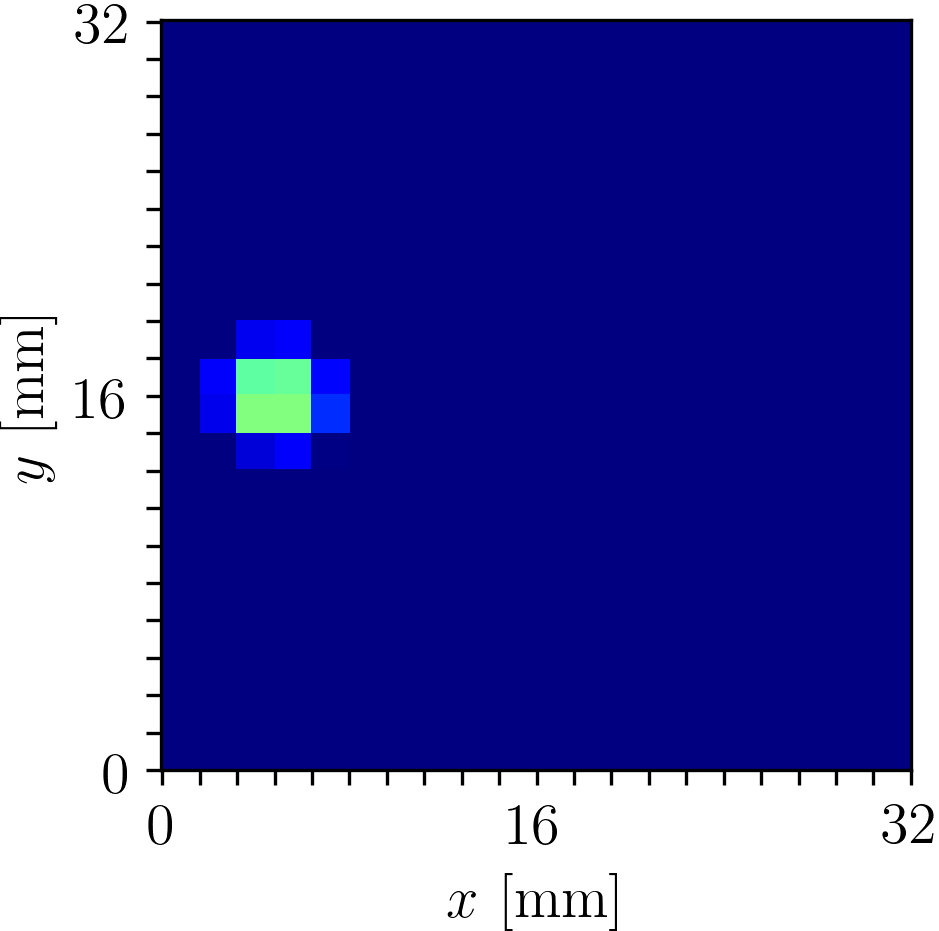}
}  \hspace{-0.6cm}\hfill
\subcaptionbox{Real-world prediction}{
	\includegraphics[height=0.45\columnwidth]{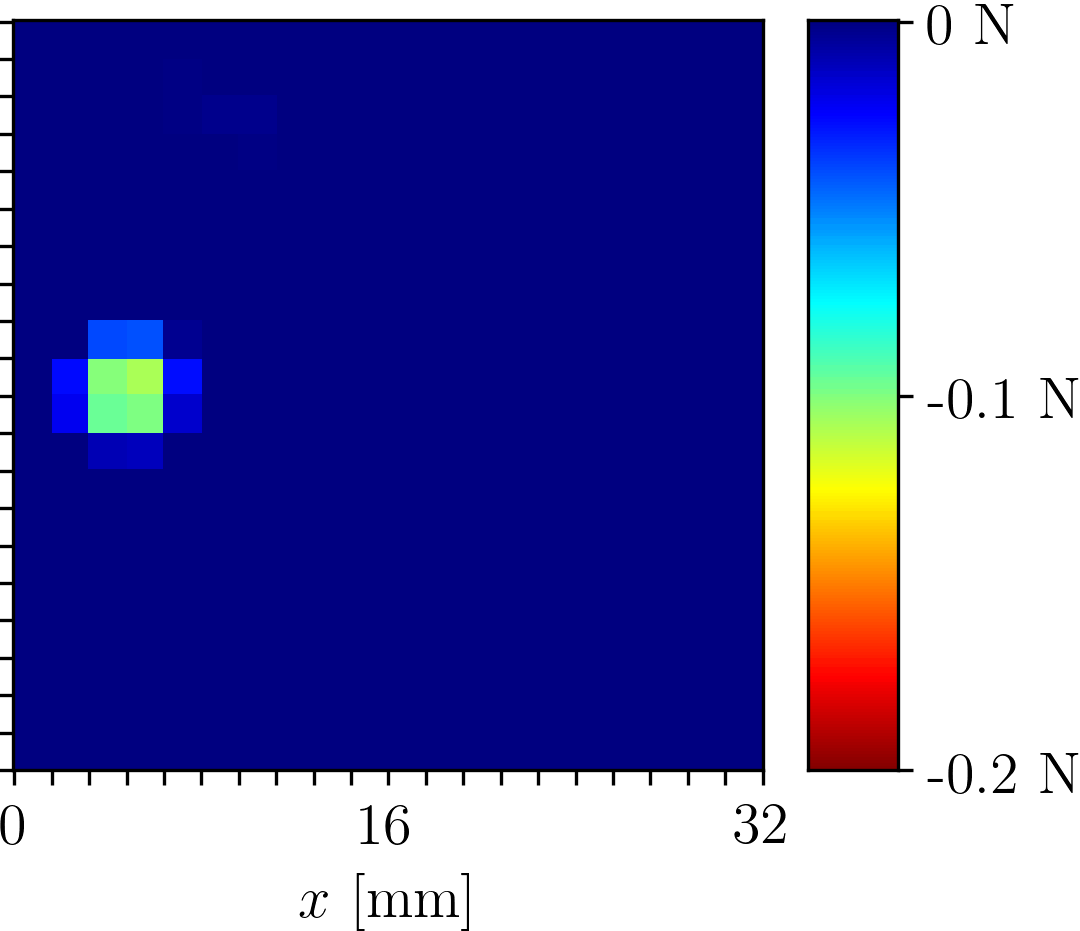}
}
	\caption{In this work, a fully synthetic dataset (see (a), (c) and (e)) is generated to train an artificial neural network, which aims to reconstruct the three-dimensional contact force distribution applied to the soft surface of a vision-based tactile sensor from optical flow (OF) features. The networks exhibits accurate predictions on real-world data (see (b), (d) and (f)). Note that in (e) and (f) only the vertical component of the contact force distribution is shown.}
	\label{fig:teaser}
	\vspace{-3mm}
\end{figure}
Understanding physical contact with the environment is a crucial requirement for the safe and reliable operation of robots interacting with their surroundings. As an example, a robot that aims to grasp objects in a cluttered box benefits from sensory feedback about the contact with these objects, in order to infer the quality of the grasp and correct its behavior \cite{calandra_more, gelslim_densebox}. Research on tactile sensing focuses on providing such feedback, generally by processing the information about the deformation of a soft sensing surface when interacting with external bodies. 

Among the various quantities of interest, the distribution of the contact forces applied to the sensing surface offers high versatility in terms of tasks and conditions. In fact, the contact force distribution retains information about the total force applied and it densely encodes the surface patches in contact with external objects. Additionally, its distributed nature provides a representation that generalizes to various contact conditions, i.e., interaction with a generic number of objects, sensing surfaces of arbitrary shape and size. However, the complexity of accurately modeling soft materials has hindered the development of sensors that can accurately reconstruct the contact force distribution in real-time, especially in the case of large deformations of the soft sensing surface. As a matter of fact, when soft materials, e.g. rubber, deform beyond a limited linear elastic region, the stress-strain relation becomes nonlinear \cite{hyperelastic_models}. As a result, accurately mapping the information about the material deformation to the contact forces generating it becomes challenging and often computationally infeasible in real-time for general cases.

In order to overcome this limitation, machine learning approaches have been proposed to obtain a data-driven model, which approximates the mapping of interest and yields appropriate inference times. A drawback in these approaches is the large amount of data typically required for each sensor produced and their limited generalization capabilities when applied to data not seen during the training time. This paper proposes a strategy to train a deep neural network with data obtained via highly accurate, state-of-the-art simulations of a vision-based tactile sensor, exploiting hyperelastic material models and an ideal camera projection. The network is directly deployed to a real-world application, where the images are transformed to the reference simulation camera model. In this way, the network does not need to be trained for each sensor, but only the appropriate camera calibration parameters need to be extracted. The real-world evaluation shows an accurate transfer from simulation to reality (sim-to-real, or sim2real), and a refinement strategy is proposed to further improve the real-world performance with a single indentation. In the case considered in this paper, the training dataset only consists of simulations of single spherical indentations. However, the model deployed in reality shows promising generalization capabilities to multiple contact conditions and to indenters of different shapes. The sensing pipeline presented here runs on the CPU of a standard laptop computer (dual-core, \mbox{2.80 GHz}) at \mbox{50 Hz}.

\subsection{Related work}
The estimation of contact forces using tactile sensors has been largely investigated in the context of contact with single objects, where the force magnitude and location, or the object shape is of interest. Several approaches have been developed for resistive \cite{weiss_resistive}, barometric measurement-based \cite{biotac_force}, capacitive \cite{capacitive_kk}, and optical \cite{gelsight_sensors} tactile sensors, either using model-based or data-driven methods. Compared to these categories, optical (or vision-based) tactile sensors exhibit very high resolution, low cost and ease of manufacture, at the expense of non-trivial data processing.

In the context of vision-based tactile sensors, the reconstruction of the contact force distribution was first \mbox{addressed} in \cite{gelforce}. The soft material was modeled as an infinite, linear elastic half-space and a closed-form solution was proposed to map the deformation of the material to the contact force distribution. In \cite{gelslim_fem}, a technique based on the finite element method (FEM) was proposed to estimate the force distribution in real-time, given the assumption of linear elasticity. The same assumption was used in \cite{softbubble_fem} to formulate an optimization-based method that estimated the surface patches in contact with external objects. While data-driven techniques are potentially suitable to overcome the complexity of modeling soft materials without introducing assumptions only valid for small deformations, their application to the estimation of distributed quantities has mainly been prevented by the lack of a ground truth source, which is crucial for supervised learning approaches. However, in \cite{sferrazza_fem} a strategy based on offline finite element simulations was recently proposed to address this problem and provide ground truth labels for generic data-driven tactile sensors. A hyperelastic model was shown to outperform a linear elastic formulation also for rather small deformations.

The major drawback of learning-based approaches to tactile sensing, either estimating total forces or distributed quantities, lies in their high data requirements. To partially address this issue, a transfer learning approach was proposed in \cite{sferrazza_transfer} to reduce the amount of data needed to transfer a learning architecture across the different sensors produced. Recently, simulation approaches have been investigated for different sensing principles in order to perform most of the training phase with synthetic data. In \cite{syntouch_sim}, a simulation model was presented to generate raw data for a sensor based on barometric measurements, with the location and magnitude of the force applied at a contact point as inputs. Tactile images for an optical sensor based on multi-color LEDs were generated in simulation in \cite{gelsight_sim} for various deformations of the sensing surface. Binary tactile contacts were simulated in the context of a grasping scenario in \cite{mat_grasping}, while a sim-to-real approach was investigated in \cite{eit_sim} for the estimation of conductivity on the surface of a tactile sensor based on electrical impedance tomography.

This work aims to provide a simulation strategy to generate an entire supervised learning dataset for a vision-based tactile sensor, with the objective of estimating the full contact force distribution from real-world tactile images. The sensor was first presented in \cite{sferrazza_sensors} and is based on a camera that tracks a random spread of particles within a soft, transparent material to infer information about the forces applied to the sensing surface. In the proposed design, all pixels of the camera provide informative data, which can be leveraged via a machine learning architecture aiming to reconstruct the three-dimensional contact force distribution with high accuracy, as shown in \cite{sferrazza_fem}. As opposed to \cite{sferrazza_fem}, where simulated ground truth labels for the force distribution were matched to real-world images, this work proposes to also generate the tactile images in simulation and to use the entire synthetic dataset in a supervised learning fashion. To this purpose, deformation data were generated via FEM simulations based on hyperelastic material models and fed through an ideal pinhole camera model. For real-world deployment, the tactile images obtained on a real tactile sensor were transformed to the pinhole reference model, by employing state-of-the-art calibration methods. In this way, not only can the model trained in simulation be deployed to real-world sensors, but it can also be easily transfered across multiple instances of the sensors produced, provided that the calibration model has been extracted. 

\subsection{Outline}
The sensing principle and the hardware used to evaluate the approach presented here are discussed in Section \ref{sec:sensing_principle}. The generation of synthetic tactile images and ground truth labels are described in Section \ref{sec:sim_learning}. In Section \ref{sec:real_data}, the transformation applied to real-world data is presented, as well as the procedure to obtain an accurate camera projection model. The results are discussed in Section \ref{sec:results}, while \mbox{Section \ref{sec:conclusion}} draws the conclusions and gives an outlook on future work. In the remainder of this paper, vectors are expressed as tuples for ease of notation, with dimension and stacking clear from the context.
\section{SENSING PRINCIPLE} \label{sec:sensing_principle}
\begin{figure}
	\centering
	\includegraphics[width=0.9\linewidth]{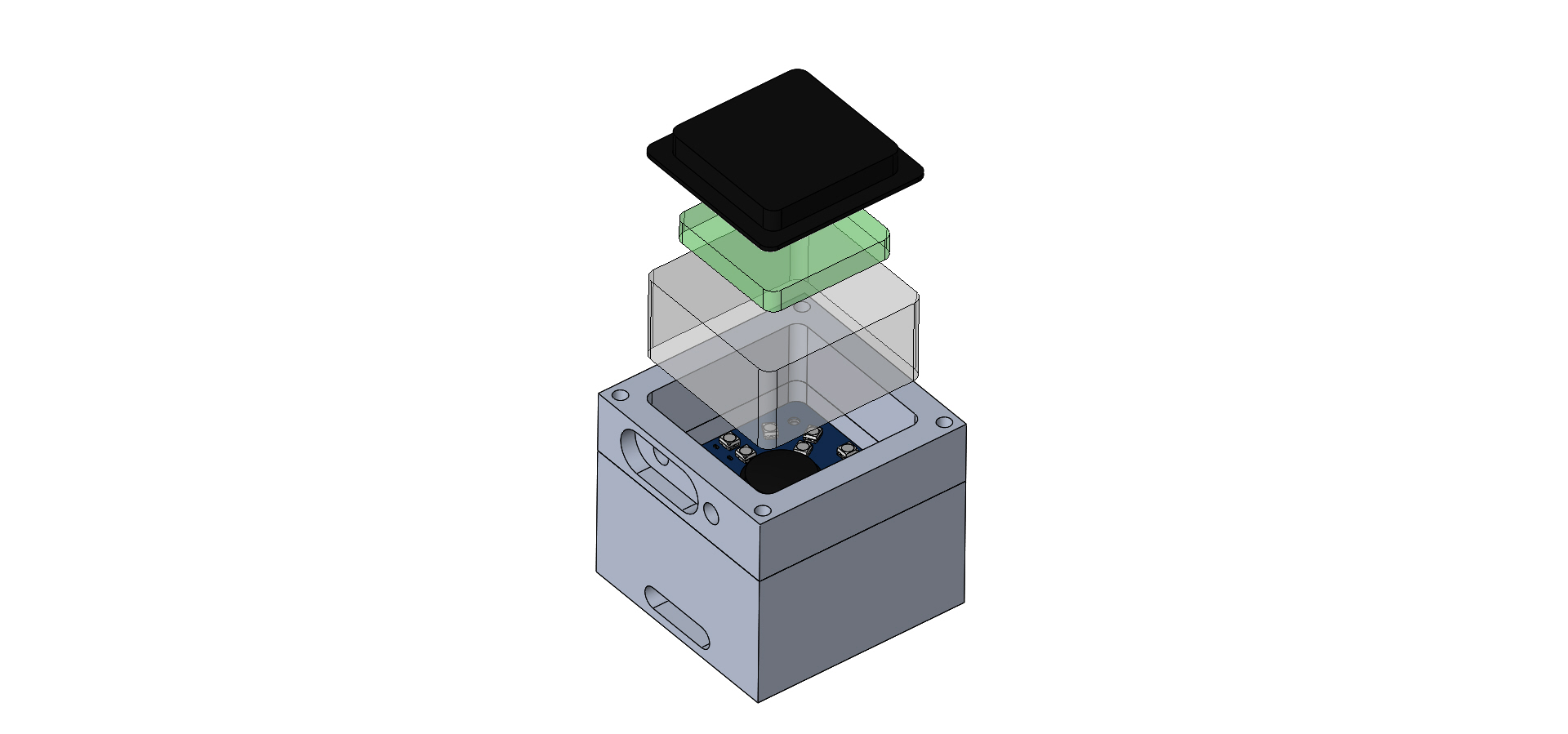}
	\caption{An exploded view of the tactile sensor employed in this paper is shown in the figure above.}
	\label{fig:exploded_view}
	\vspace{-5mm}
\end{figure}

The sensor employed in this paper follows the principle introduced in \cite{sferrazza_sensors}. Three soft silicone layers are poured on top of an RGB fisheye camera (ELP USBFHD06H), surrounded by LEDs. From the bottom: a stiff transparent layer (ELASTOSIL\textregistered{} RT 601 RTV-2, mixing  ratio 7:1, shore hardness 45A), which serves as a spacer and for light diffusion; a very soft transparent layer (Ecoflex\textsuperscript{TM} GEL, ratio 1:1, shore hardness 000-35), which embeds a spread of randomly distributed polyethylene particles (microspheres with a diameter of 150 to 180 $\mu$m); a black layer (ELASTOSIL\textregistered{} RT 601 RTV-2, ratio 25:1, shore hardness 10A), which is more resistant to repeated contact than the middle layer and shields the sensor from external light. An exploded view of the sensor layers is depicted in Fig.~\ref{fig:exploded_view}. The volume of the gel containing the particles is 30$\times$30$\times$4.5 mm, while the joint volume of the particle layer and the black layer is \mbox{32$\times$32$\times$6 mm}.

When the soft sensing surface is subject to force, the material deforms and displaces the particles tracked by the camera. This motion generates different patterns in the images, which can be processed to extract information about the contact force distribution causing the deformation.

\section{LEARNING IN SIMULATION} \label{sec:sim_learning}
\begin{figure}[]
	\centering
	\subcaptionbox{Dataset generation as in \cite{sferrazza_fem}}{
	\scalebox{0.25}{\definecolor{red}{rgb}{1,0,0}%
\definecolor{green}{rgb}{0,1,0}%
\definecolor{blue}{rgb}{0,0,1}%
\definecolor{yellow}{rgb}{1,1,0}%
\definecolor{orange}{rgb}{1,0.647,0}%
\definecolor{gold}{rgb}{1,0.843,0}%
\definecolor{purple}{rgb}{0.627,0.125,0.941}%
\definecolor{gray}{rgb}{0.745,0.745,0.745}%
\definecolor{brown}{rgb}{0.647,0.165,0.165}%
\definecolor{navy}{rgb}{0,0,0.502}%
\definecolor{pink}{rgb}{1,0.753,0.796}%
\definecolor{seagreen}{rgb}{0.18,0.545,0.341}%
\definecolor{turquoise}{rgb}{0.251,0.878,0.816}%
\definecolor{violet}{rgb}{0.933,0.51,0.933}%
\definecolor{darkblue}{rgb}{0,0,0.545}%
\definecolor{darkcyan}{rgb}{0,0.545,0.545}%
\definecolor{darkgray}{rgb}{0.663,0.663,0.663}%
\definecolor{darkgreen}{rgb}{0,0.392,0}%
\definecolor{darkmagenta}{rgb}{0.545,0,0.545}%
\definecolor{darkorange}{rgb}{1,0.549,0}%
\definecolor{darkred}{rgb}{0.545,0,0}%
\definecolor{lightblue}{rgb}{0.678,0.847,0.902}%
\definecolor{lightcyan}{rgb}{0.878,1,1}%
\definecolor{lightgray}{rgb}{0.827,0.827,0.827}%
\definecolor{verylightgray}{rgb}{0.927,0.927,0.927}%
\definecolor{lightgreen}{rgb}{0.565,0.933,0.565}%
\definecolor{customgreen}{rgb}{0.796,0.890,0.824}%
\definecolor{lightyellow}{rgb}{1,1,0.878}%
\definecolor{black}{rgb}{0,0,0}%
\definecolor{white}{rgb}{1,1,1}%
\newcommand{\HUGE}{\fontsize{35}{60}\selectfont}%
\newcommand{\HHUGE}{\fontsize{45}{60}\selectfont}%
\renewcommand{\arraystretch}{3.5}%
\begin{tikzpicture}[every text node part/.style={align=center}]
  \node at (0,0) [draw, line width = 1.5mm,rounded corners, fill=verylightgray, minimum width = 13cm, minimum height = 10cm] (R1) {};%
  \node [below = of R1.north] (I1) {\includegraphics[width=0.55\textwidth]{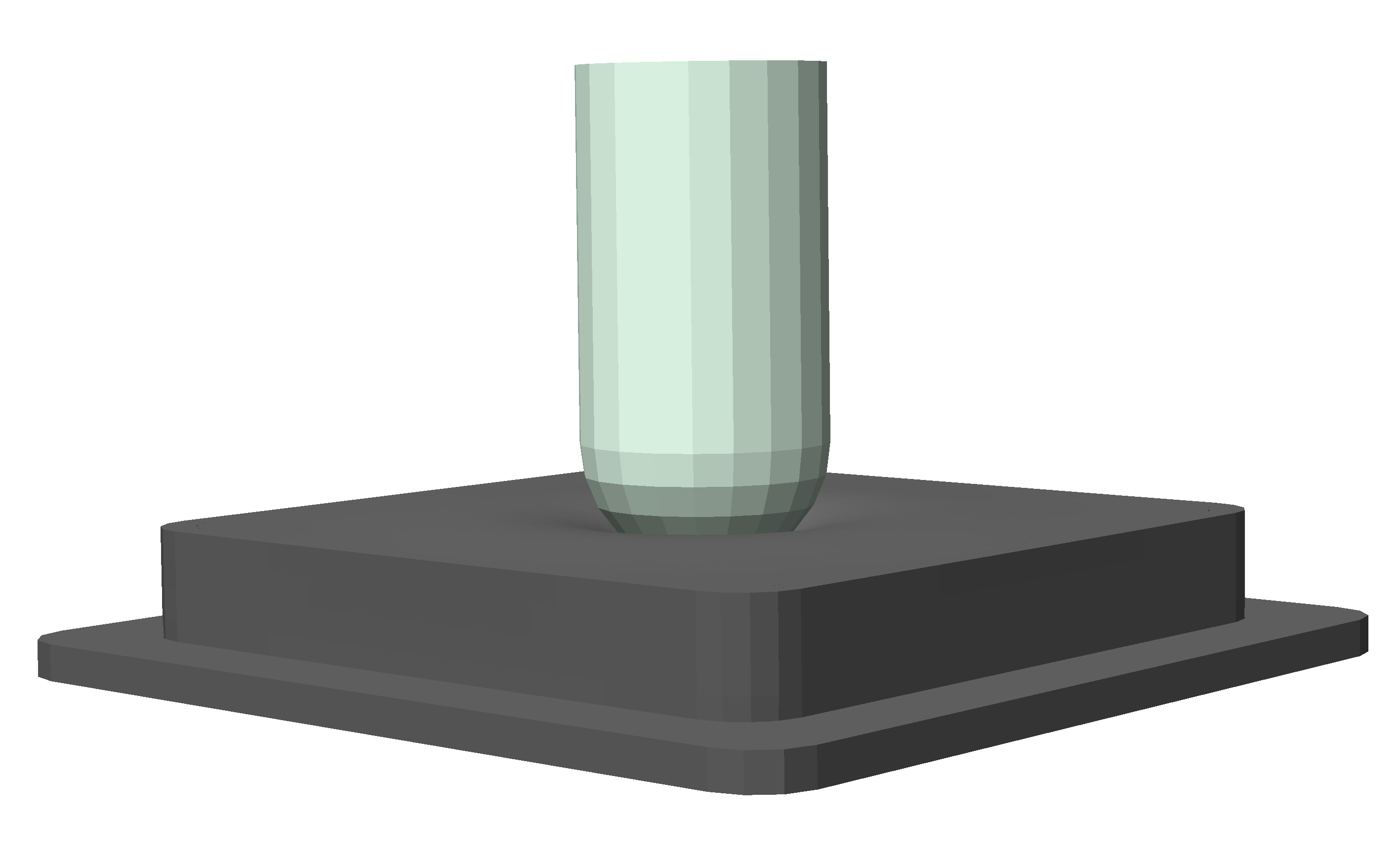}};%
  \node[below = of I1.south, yshift = 2mm] {\HUGE{FEM simulations}};%
  \coordinate [below = of R1] (C);%
  \node  [below = of R1.south ,draw, line width = 1.5mm,rounded corners, fill=verylightgray, minimum width = 13cm, minimum height = 10cm] (R2) {};%
  \node [below = of R2.north, yshift = 0cm ] (I2) {\includegraphics[width=0.55\textwidth]{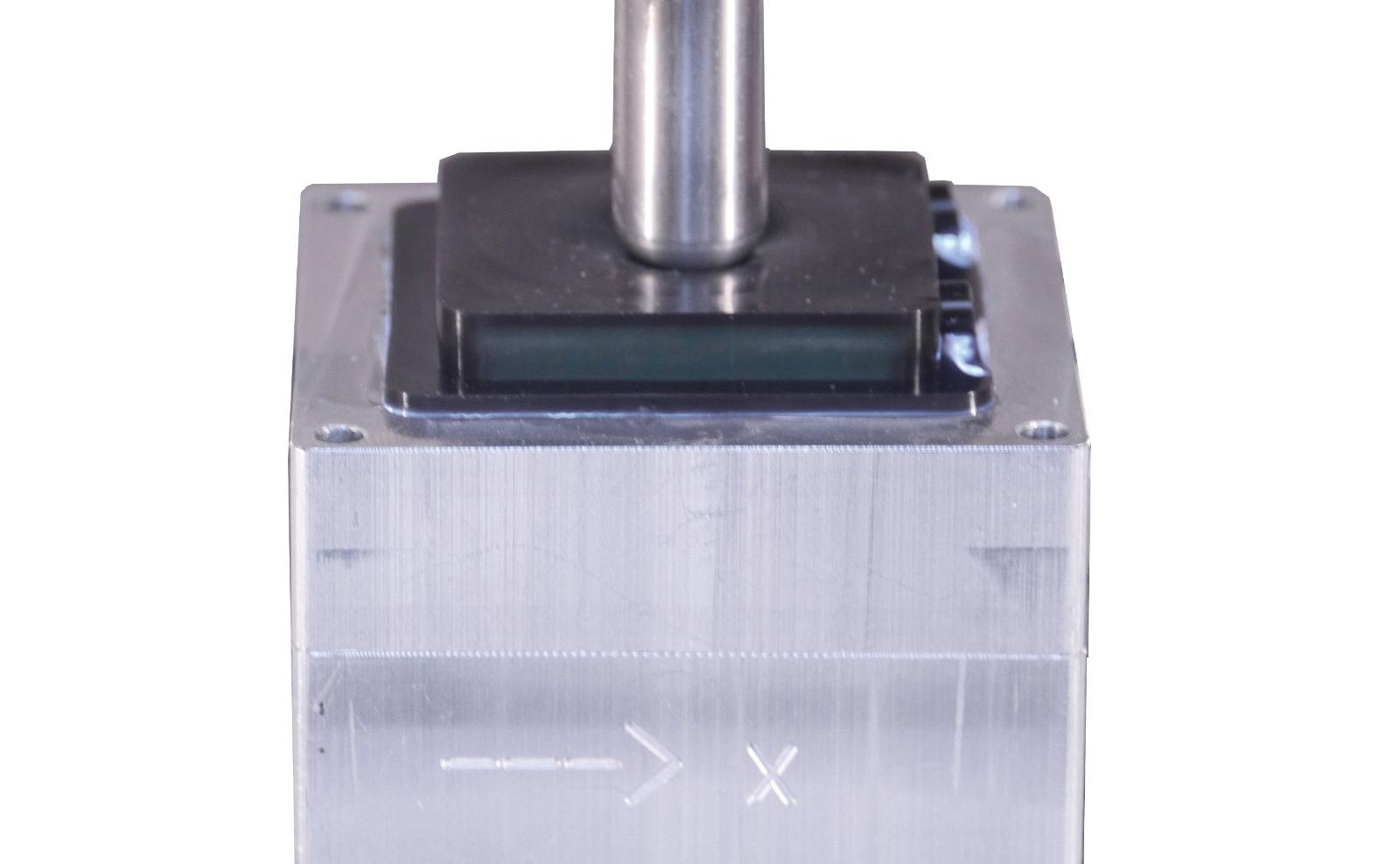}};%
  \node[below =of I2.south, yshift = 0mm] {\HUGE{Real-world indentations}};%
  \node [right = of C.east,yshift = 5mm, xshift = 15cm, draw, line width = 1.5mm,rounded corners, fill=customgreen, minimum width = 10cm, minimum height = 10cm, text width = 8 cm] (R3) {\begin{tabular}{c} \HHUGE{Training} \\ \HHUGE{+} \\ \HHUGE{Evaluation} \end{tabular}};%
  \coordinate [below = of R1.south, anchor = north, yshift=  3cm] (D);
  \coordinate [below = of R1.south, anchor = north, yshift=  -2cm] (E);
  \draw [-{Latex[length=8mm]}, thick, line width = 1.5mm] (D-|R1.east) -- (D-|R3.west) node[midway,above] {\begin{tabular}{c} \HUGE{Ground truth} \\ \HUGE{force distribution} \end{tabular}};
  \draw [-{Latex[length=8mm]}, thick, line width = 1.5mm] (E-|R2.east) -- (E-|R3.west) node[midway,above] {\HUGE{DIS optical flow}};
\end{tikzpicture}}
	} \\
\vskip\baselineskip
	\subcaptionbox{Dataset generation proposed here}{
	\scalebox{0.25}{\input{img/synthetic_training.tex}}
	} \hfill
	\caption{As shown in (a), the strategy proposed in \cite{sferrazza_fem} was based on the collection of real-world images, from which optical flow features were extracted. Ground truth labels were obtained in simulation and assigned to these features to train a supervised learning architecture. The resulting architecture was then evaluated on a portion of this dataset, not used during training. Here the training dataset is fully generated in simulation by extracting synthetic optical flow features from FEM indentations. In this step, a reference pinhole camera model is employed. The sim-to-real transfer is evaluated on a dataset composed of real optical flow features, which are computed after the real-world (distorted) images are remapped to the reference pinhole model.}
	\label{fig:flow_diagram}
	\vspace{-6mm}
\end{figure}

The contact force distribution is modeled here in a discretized fashion, by dividing the square sensing surface in \mbox{$n \times n$} bins of equal area. Three matrices $F^G_x, F^G_y, F^G_z$ of size \mbox{$n \times n$} represent the force distribution in the gel coordinate system, where the origin is placed at one of the surface corners, $x^G$ and $y^G$ are aligned with two perpendicular surface edges and $z^G$ is the vertical axis, pointing from the camera towards the surface, as shown in Fig.~\ref{fig:frames_definition}. Each matrix element represents the respective force component applied at the respective bin.

The reconstruction of the force distribution can be posed as a supervised learning problem, even if there are no readily available commercial sensors that can measure the ground truth, i.e., the three-dimensional contact force distribution applied to soft materials, without altering the sensing surface. In fact, in \cite{sferrazza_fem} it was shown how highly accurate ground truth force distributions can be obtained by means of FEM simulations, where the corresponding nodal forces are summed within each bin to obtain the force distribution matrices. In \cite{sferrazza_fem}, hyperelastic models of the soft materials employed were obtained via state-of-the-art characterization methods and used to simulate thousands of indentation experiments. In order to create a supervised learning dataset (see Fig.~\ref{fig:flow_diagram}), the resulting ground truth labels were then matched to optical flow features extracted via an algorithm based on dense inverse search (DIS \cite{dis_paper}). These features were obtained by replicating in reality the same indentation experiments. Conversely, in this paper the optical flow features, which are the input to the learning algorithm, are obtained as well from the simulated indentations and together with the ground truth labels contribute to generating a fully synthetic training dataset, greatly reducing the data collection efforts. The sim-to-real transfer is evaluated on real-world indentations performed at sampled surface locations, as discussed in Section \ref{sec:real_data}.

\subsection{Generating optical flow features}
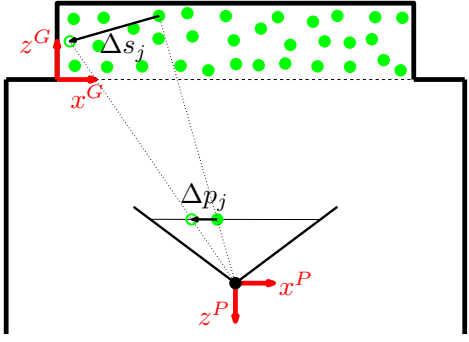
\begin{figure}
	%\centering
	\scalebox{0.3}{\tikzstyle{ipe stylesheet} = [
  ipe import,
  even odd rule,
  line join=round,
  line cap=butt,
  ipe pen normal/.style={line width=1},
  ipe pen heavier/.style={line width=2},
  ipe pen fat/.style={line width=3},
  ipe pen ultrafat/.style={line width=6},
  ipe pen normal,
  ipe mark normal/.style={ipe mark scale=3},
  ipe mark large/.style={ipe mark scale=13},
  ipe mark small/.style={ipe mark scale=2},
  ipe mark tiny/.style={ipe mark scale=1.1},
  ipe mark normal,
  /pgf/arrow keys/.cd,
  ipe arrow normal/.style={scale=7},
  ipe arrow large/.style={scale=10},
  ipe arrow small/.style={scale=5},
  ipe arrow tiny/.style={scale=3},
  ipe arrow normal,
  /tikz/.cd,
  ipe arrows, % update arrows
  <->/.tip = ipe normal,
  ipe dash normal/.style={dash pattern=},
  ipe dash dashed/.style={dash pattern=on 4bp off 4bp},
  ipe dash dotted/.style={dash pattern=on 1bp off 3bp},
  ipe dash dash dotted/.style={dash pattern=on 4bp off 2bp on 1bp off 2bp},
  ipe dash dash dot dotted/.style={dash pattern=on 4bp off 2bp on 1bp off 2bp on 1bp off 2bp},
  ipe dash normal,
  ipe node/.append style={font=\normalsize},
  ipe stretch normal/.style={ipe node stretch=1},
  ipe stretch normal,
  ipe opacity 10/.style={opacity=0.1},
  ipe opacity 30/.style={opacity=0.3},
  ipe opacity 50/.style={opacity=0.5},
  ipe opacity 75/.style={opacity=0.75},
  ipe opacity opaque/.style={opacity=1},
  ipe opacity opaque,
]
\definecolor{red}{rgb}{1,0,0}
\definecolor{green}{rgb}{0,1,0}
%\definecolor{green}{rgb}{0.796,0.890,0.824}%
\definecolor{blue}{rgb}{0,0,1}
\definecolor{yellow}{rgb}{1,1,0}
\definecolor{orange}{rgb}{1,0.647,0}
\definecolor{gold}{rgb}{1,0.843,0}
\definecolor{purple}{rgb}{0.627,0.125,0.941}
\definecolor{gray}{rgb}{0.745,0.745,0.745}
\definecolor{brown}{rgb}{0.647,0.165,0.165}
\definecolor{navy}{rgb}{0,0,0.502}
\definecolor{pink}{rgb}{1,0.753,0.796}
\definecolor{seagreen}{rgb}{0.18,0.545,0.341}
\definecolor{turquoise}{rgb}{0.251,0.878,0.816}
\definecolor{violet}{rgb}{0.933,0.51,0.933}
\definecolor{darkblue}{rgb}{0,0,0.545}
\definecolor{darkcyan}{rgb}{0,0.545,0.545}
\definecolor{darkgray}{rgb}{0.663,0.663,0.663}
\definecolor{darkgreen}{rgb}{0,0.392,0}
\definecolor{darkmagenta}{rgb}{0.545,0,0.545}
\definecolor{darkorange}{rgb}{1,0.549,0}
\definecolor{darkred}{rgb}{0.545,0,0}
\definecolor{lightblue}{rgb}{0.678,0.847,0.902}
\definecolor{lightcyan}{rgb}{0.878,1,1}
\definecolor{lightgray}{rgb}{0.827,0.827,0.827}
\definecolor{lightgreen}{rgb}{0.565,0.933,0.565}
\definecolor{lightyellow}{rgb}{1,1,0.878}
\definecolor{black}{rgb}{0,0,0}
\definecolor{white}{rgb}{1,1,1}
\newcommand{\HUGE}{\fontsize{35}{60}\selectfont}%
\newcommand{\HHUGE}{\fontsize{45}{60}\selectfont}%
\begin{tikzpicture}[ipe stylesheet]
  \pic[ipe mark large, green]
     at (272, 496) {ipe circle};
  \draw[ipe pen ultrafat]
    (192, 128)
     -- (192, 448);
  \draw[ipe pen ultrafat]
    (192, 448)
     -- (256, 448)
     -- (256, 448);
  \draw[ipe pen ultrafat]
    (256, 448)
     -- (256, 544)
     -- (704, 544)
     -- (704, 448);
  \draw[ipe pen ultrafat]
    (704, 448)
     -- (768, 448)
     -- (768, 448);
  \draw[ipe pen ultrafat]
    (768, 448)
     -- (768, 128);
  \draw[ipe dash dashed]
    (256, 448)
     -- (704, 448);
  \pic[ipe mark large, green]
     at (276.808, 524.2324) {ipe disk};
  \pic[ipe mark large, green]
     at (278.2292, 465.2516) {ipe disk};
  \pic[ipe mark large, green]
     at (362.7921, 464.541) {ipe disk};
  \pic[ipe mark large, green]
     at (318.7341, 463.8304) {ipe disk};
  \pic[ipe mark large, green]
     at (383.3998, 499.361) {ipe disk};
  \pic[ipe mark large, green]
     at (426.0365, 502.2034) {ipe disk};
  \pic[ipe mark large, green]
     at (445.9337, 459.5667) {ipe disk};
  \pic[ipe mark large, green]
     at (480.7537, 467.3834) {ipe disk};
  \pic[ipe mark large, green]
     at (509.1782, 465.9622) {ipe disk};
  \pic[ipe mark large, green]
     at (542.5769, 468.0941) {ipe disk};
  \pic[ipe mark large, green]
     at (477.9112, 497.2292) {ipe disk};
  \pic[ipe mark large, green]
     at (533.339, 526.3643) {ipe disk};
  \pic[ipe mark large, green]
     at (548.9724, 491.5443) {ipe disk};
  \pic[ipe mark large, green]
     at (577.3969, 520.6794) {ipe disk};
  \pic[ipe mark large, green]
     at (583.0818, 464.541) {ipe disk};
  \pic[ipe mark large, green]
     at (623.5867, 494.3867) {ipe disk};
  \pic[ipe mark large, green]
     at (651.3006, 498.6504) {ipe disk};
  \pic[ipe mark large, green]
     at (655.5643, 526.3643) {ipe disk};
  \pic[ipe mark large, green]
     at (689.6737, 525.6536) {ipe disk};
  \pic[ipe mark large, green]
     at (666.2235, 472.3577) {ipe disk};
  \pic[ipe mark large, green]
     at (687.5418, 459.5667) {ipe disk};
  \pic[ipe mark large, green]
     at (419.641, 531.3385) {ipe disk};
  \pic[ipe mark large, green]
     at (467.9627, 527.7855) {ipe disk};
  \pic[ipe mark large, green]
     at (411.1137, 465.2516) {ipe disk};
  \pic[ipe mark large, green]
     at (314.4704, 525.6536) {ipe disk};
  \pic[ipe mark large, green]
     at (308.0749, 490.123) {ipe disk};
  \pic[ipe mark large, green]
     at (384, 528) {ipe disk};
  \draw[ipe pen fat]
    (480, 192)
     -- (352, 288);
  \draw[ipe pen fat]
    (480, 192)
     -- (608, 288);
  \draw
    (372.9707, 272.2719)
     -- (586.6296, 271.9722)
     -- (586.857, 272.1427);
  \pic[ipe mark large, green]
     at (457.1314, 272.1539) {ipe disk};
  \draw[red, ipe pen ultrafat, ->]
    (256, 448)
     -- node [left,pos=1] { \scalebox{3.5}{$z^G$}} (256, 496);
  \draw[red, ipe pen ultrafat, ->]
    (256, 448)
     -- node [below,pos=0.8]{ \scalebox{3.5}{$x^G$}} (304, 448);
  \draw[red, ipe pen ultrafat, ->]
    (480, 192)
     -- node [left,pos=0.8]{ \scalebox{3.5}{$z^P$}} (480, 144);
  \draw[red, ipe pen ultrafat, ->]
    (480, 192)
     -- node [right,pos=1]{ \scalebox{3.5}{$x^P$}}(528, 192);
  \pic[ipe mark large, green]
     at (512, 512) {ipe disk};
  \pic[ipe mark large, green]
     at (592, 496) {ipe disk};
  \pic[ipe mark large, green]
     at (608, 528) {ipe disk};
  \pic[ipe mark large, green]
     at (624, 464) {ipe disk};
  \pic[ipe mark large, green]
     at (352, 512) {ipe disk};
  \draw[ipe pen fat, ->]
    (384, 528)
     -- node [below, pos=0.4, yshift=-2mm]{ \scalebox{3.5}{$\Delta s_j$}}(272, 496);
  \draw[ipe dash dotted]
    (480, 192)
     -- (384, 528);
  \draw[ipe dash dotted]
    (480, 192)
     -- (272, 496);
  \pic[ipe mark large, green]
     at (424.9245, 272.199) {ipe circle};
  \draw[ipe pen fat, ->]
    (456.8352, 272.1543)
     --  node [above, pos=0.5, yshift = 2mm]{ \scalebox{3.5}{$\Delta p_j$}}(425.3481, 272.1985);
     \pic[ipe mark large]
     at (480, 192) {ipe disk};
\end{tikzpicture}}
	\caption{The figure depicts the coordinate systems employed, here shown in the two-dimensional case. The displacement vectors are retrieved from the FEM simulations in the gel coordinate system (superscript $G$) and transformed to the pinhole coordinate system (superscript $P$). Finally, the projected pixel displacements are computed as described in \eqref{eq:disp_projection2}-\eqref{eq:pos_projection2}.}
	\label{fig:frames_definition}\vspace{-3mm}
\end{figure}
The DIS optical flow algorithm estimates the displacement of the particles at each pixel between a reference frame (with the gel at rest) and the current frame. In order to replicate this functionality on the simulated indentations, the undeformed gel volume containing the particles is first sampled on a fine uniform grid of points. In the remainder of this paper, the positions of these points with the gel at rest are denoted as \textit{undeformed locations}, while their positions after deformation are denoted as \textit{deformed locations}. For each indentation, the respective displacement vectors are assigned to the undeformed locations, via an inverse distance weighted interpolation \cite{spatial_interpolation} of the displacement field obtained from the FEM simulations. This step ensures that 3D displacement vectors are estimated at uniformly spaced locations, since the FEM mesh is generally refined at sections of interest. Each of these displacement vectors emulates the motion that a particle located at the respective undeformed location at rest would experience during an indentation. Within a specific indentation, this means that a particle with an initial location in the gel coordinate system at 
\begin{align}
s^G_j := (x^G_j,y^G_j,z^G_j)
\end{align}
is expected to move to a deformed location $s^G_j +  \Delta s^G_j$,
where
\begin{align}
\Delta s^G_j:= (\Delta x^S_j,\Delta y^S_j,\Delta z^S_j)
\end{align} 
represents the corresponding displacement vector, for \mbox{$j = 0,\dots,N_\text{s}-1$}, with $N_\text{s}$ the number of sampled locations. In order to simulate an optical flow estimation, the displacement vectors are projected to the image plane. This involves a coordinate transformation, from the gel coordinate system (aligned with the simulation coordinate system) to the image, employing an ideal pinhole camera model \cite[p.~49]{computervision_book}, as depicted in Fig.~\ref{fig:frames_definition}. To this purpose, a displacement vector and its corresponding undeformed location are first transformed to the 3D pinhole camera coordinate system as
\begin{gather}
\Delta s_j^P = R^{GP} \Delta s_j^G, \\
s_j^P  = R^{GP} s_j^G + t^{GP}, \label{eq:pos_projection1}
\end{gather}
where $R^{GP}$ and $t^{GP} := (t^{GP}_x,t^{GP}_y,t^{GP}_z)$ are the corresponding rotation matrix and a translation vector, respectively, comprising the reference camera's extrinsic parameters. The choice of these parameters is further discussed in \mbox{Section \ref{sec:real_data}}. The resulting pixel displacement and the projection of the corresponding undeformed location on the image, i.e., $\Delta p_j := (\Delta u_j, \Delta v_j)$ and $p_j : = (u_j,v_j)$, respectively, are then computed as
\begin{gather}\label{eq:disp_projection2}
\Delta p_j = K_{\text{a},j}(s^P_j + \Delta s^P_j) - K_{\text{b},j}s^P_j,\\
p_j = K_{\text{b},j}s^P_j \label{eq:pos_projection2},
\end{gather}
with
\begin{gather}
K_{\text{a},j} = \dfrac{1}{z^P_j + \Delta z^P_j}\begin{bmatrix}
f & 0 & u_\text{c} \\
0 & f & v_\text{c} 
\end{bmatrix}, \\
K_{\text{b},j} = \dfrac{1}{z^P_j}\begin{bmatrix}
f& 0 & u_\text{c} \\
0 & f & v_\text{c}
\end{bmatrix},
\end{gather}
where $|f|$ is the focal length and the camera center coordinates $u_\text{c}, v_\text{c}$ are set at the image center. Since the particle layer has a square horizontal section of 30$\times$30 mm, the image region of interest is set as a square of (arbitrary) dimension 440$\times$440 pixels and the focal length is equally chosen for both the image coordinates as 
\begin{align}
f := \dfrac{440}{30}t^{GP}_z,
\end{align}
to exactly fill the image with the particle layer. Note that $f$ is negative, due to the definition of the pinhole camera coordinate system.

In order to create a compact set of features, the image is divided into $m \times m$ regions of equal area. The pixel displacements are assigned to the image regions, based on the coordinates of the corresponding $p_j$. Then, the average of these displacements within each image region $(i,l)$, for $i=0,\dots,m-1$ and $l=0,\dots,m-1$, is computed for both components as
\begin{gather} \label{eq:average}
\overline{\Delta u_{il}} = \frac{1}{\|w_{il}\|} \sum_{j \in \mathcal{J}_{il}} w_{j}\Delta u_{j}, \\\label{eq:average2}
\overline{\Delta v_{il}} = \frac{1}{\|w_{il}\|} \sum_{j \in \mathcal{J}_{il}} w_{j}\Delta v_{j},
\end{gather}
where $\mathcal{J}_{il} \subseteq \{0,\dots,N_\text{s}-1\}$ is the set of the displacement indices assigned to the region $(i,l)$, $w_{j}$ are averaging weights and $\|w_{il}\|:= \sum_{j \in \mathcal{J}_{il}} w_j$. 

The weights are introduced to account for occlusions occurring in real-world images. In fact, since the synthetic optical flow emulates the displacement of the particles in reality, one must consider the fact that on real images some of the particles might be occluded by the particles closer to the camera. For this reason, each projected displacement is weighted in \eqref{eq:average}-\eqref{eq:average2} with the probability of both its deformed and undeformed locations being visible in the image, that is,
\begin{align}
w_{j} := \rho_{j} \sigma_{j},
\end{align}
where $\rho_{j}$ is the probability that a particle located (in the gel coordinate system) at $s_j^G$ is visible in the image frame taken with the gel at rest, and $\sigma_{j}$ is the probability that a particle located at $s_j^G + \Delta s_j^G$ is visible in the image frame taken after deformation, that is, at the time the optical flow is being computed. Note that here the two visibility events in the respective frames have been considered to be independent, which might not be a valid assumption for some special cases (e.g., for particles placed on the camera optical axis during centered vertical indentations). However, this simplifies the derivation and has proved to be a reasonable assumption in practice. The two probability values for each weight can be computed via Monte Carlo simulations, assuming that the density of the particles does not considerably change during an indentation. This is done by randomly drawing 100 different particle configurations and projecting them to the image plane using \eqref{eq:pos_projection1} and \eqref{eq:pos_projection2}. Assuming that a spherical particle is projected to a circle in the image, the radius $r$ of the projected circle is computed as shown in the Appendix, and a particle is considered as occluded if its center in the image is covered by any other circle generated by a particle closer to the camera. Note that in the calculation of the weights, the particle configurations are sampled using the known particle-to-silicone ratio. The resulting particles are generally less than $N_\text{s}$, which is chosen to be large enough to enhance robustness to numerical noise in the FEM results.

The probability of a visible particle is approximated in discrete 3D bins, to which particles are assigned depending on their position within the gel, dividing the number of visible particles in the bin by the total number of particles assigned to the respective bin. This is done for each particle layer configuration and the resulting probabilities are averaged over the 100 configurations. Both the values $\rho_{j}$ and $\sigma_j$ are retrieved from this probability discretization, depending on the locations $s_j^G$ and $s_j^G + \Delta s_j^G$ for the $j$-th displacement vector. Note that this is a valid procedure for computing $\sigma_j$ only if the density of the particles is constant over an indentation. This assumption is justified by the large number of particles spread within the gel at varying depths. However, this approximation becomes more severe for large deformations, when the particles tend not to spread homogeneously.

\subsection{Learning architecture}
The learning task is addressed here as an image-to-image translation, also known as pixel-wise regression. In fact, the quantities obtained from \eqref{eq:average}-\eqref{eq:average2} are rearranged to form an image-like tensor with two channels each with $m\times m$ elements. Similarly, the three matrices $F^G_x, F^G_y, F^G_z$ representing the force distribution are grouped in three channels each with $n\times n$ elements. In the following, the case subject of evaluation, i.e., $m=40$, $n=20$, will be considered.

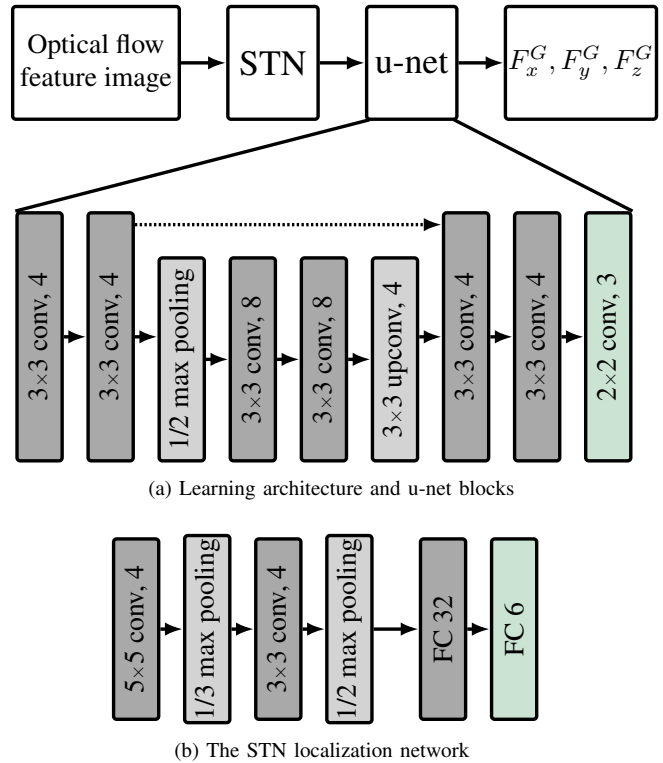
\begin{figure}
	\centering
	\subcaptionbox{Learning architecture and u-net blocks}{
	\scalebox{0.3}{\definecolor{red}{rgb}{1,0,0}%
\definecolor{green}{rgb}{0,1,0}%
\definecolor{blue}{rgb}{0,0,1}%
\definecolor{yellow}{rgb}{1,1,0}%
\definecolor{orange}{rgb}{1,0.647,0}%
\definecolor{gold}{rgb}{1,0.843,0}%
\definecolor{purple}{rgb}{0.627,0.125,0.941}%
\definecolor{gray}{rgb}{0.745,0.745,0.745}%
\definecolor{brown}{rgb}{0.647,0.165,0.165}%
\definecolor{navy}{rgb}{0,0,0.502}%
\definecolor{pink}{rgb}{1,0.753,0.796}%
\definecolor{seagreen}{rgb}{0.18,0.545,0.341}%
\definecolor{turquoise}{rgb}{0.251,0.878,0.816}%
\definecolor{violet}{rgb}{0.933,0.51,0.933}%
\definecolor{darkblue}{rgb}{0,0,0.545}%
\definecolor{darkcyan}{rgb}{0,0.545,0.545}%
\definecolor{darkgray}{rgb}{0.663,0.663,0.663}%
\definecolor{darkgreen}{rgb}{0,0.392,0}%
\definecolor{darkmagenta}{rgb}{0.545,0,0.545}%
\definecolor{darkorange}{rgb}{1,0.549,0}%
\definecolor{darkred}{rgb}{0.545,0,0}%
\definecolor{lightblue}{rgb}{0.678,0.847,0.902}%
\definecolor{lightcyan}{rgb}{0.878,1,1}%
\definecolor{lightgray}{rgb}{0.827,0.827,0.827}%
\definecolor{verylightgray}{rgb}{0.927,0.927,0.927}%
\definecolor{lightgreen}{rgb}{0.565,0.933,0.565}%
\definecolor{customgreen}{rgb}{0.796,0.890,0.824}%
\definecolor{lightyellow}{rgb}{1,1,0.878}%
\definecolor{black}{rgb}{0,0,0}%
\definecolor{white}{rgb}{1,1,1}%
\newcommand{\HUGE}{\fontsize{35}{60}\selectfont}%
\newcommand{\HHUGE}{\fontsize{45}{60}\selectfont}%
\renewcommand{\arraystretch}{3.5}%
\begin{tikzpicture}[every text node part/.style={align=center}]
\node at (0,0) [draw, line width = 1.5mm,rounded corners, minimum width = 6cm, minimum height = 5cm] (R_inp) {\begin{tabular}{c} \HUGE{Optical flow} \\ \HUGE{feature image} \end{tabular}};%

\node [right = of R_inp.east, draw, line width = 1.5mm,rounded corners, minimum width = 4cm, minimum height = 5cm, xshift=1cm] (R_stn) {\HHUGE STN};%

\node [right = of R_stn.east, draw, line width = 1.5mm,rounded corners, minimum width = 4cm, minimum height = 5cm, xshift=1cm] (R_unet) {\HHUGE u-net};%

\node [right = of R_unet.east, draw, line width = 1.5mm,rounded corners, minimum width = 6cm, minimum height = 5cm, xshift=1cm] (R_out) {\scalebox{3.5}{$F_x^G,F_y^G,F_z^G$}};%

\draw [-{Latex[length=8mm]}, thick, line width = 1.5mm] (R_inp.east) -- (R_stn.west);
\draw [-{Latex[length=8mm]}, thick, line width = 1.5mm] (R_stn.east) -- (R_unet.west);
\draw [-{Latex[length=8mm]}, thick, line width = 1.5mm] (R_unet.east) -- (R_out.west);

\node [below = of R_inp.south,draw, line width = 1.5mm,rounded corners, minimum width = 2cm, minimum height = 11cm, xshift = -2.5cm, yshift = -3cm, fill = darkgray] (R_conv1){\rotatebox{90}{\HUGE 3$\times$3 conv, 4}};%
\node [right = of R_conv1.east,draw, line width = 1.5mm,rounded corners, minimum width = 2cm, minimum height = 11cm, fill = darkgray] (R_conv2){\rotatebox{90}{\HUGE 3$\times$3 conv, 4}};%
\node [right = of R_conv2.east,draw, line width = 1.5mm,rounded corners, minimum width = 2cm, minimum height = 9cm, fill = lightgray, yshift = -1cm] (R_mp){\rotatebox{90}{\HUGE 1/2 max pooling}};%
\node [right = of R_mp.east,draw, line width = 1.5mm,rounded corners, minimum width = 2cm, minimum height = 9cm, fill = darkgray] (R_conv3){\rotatebox{90}{\HUGE 3$\times$3 conv, 8}};%
\node [right = of R_conv3.east,draw, line width = 1.5mm,rounded corners, minimum width = 2cm, minimum height = 9cm, fill = darkgray] (R_conv4){\rotatebox{90}{\HUGE 3$\times$3 conv, 8}};%
\node [right = of R_conv4.east,draw, line width = 1.5mm,rounded corners, minimum width = 2cm, minimum height = 9cm, fill = lightgray] (R_up){\rotatebox{90}{\HUGE 3$\times$3 upconv, 4}};%
\node [right = of R_up.east,draw, line width = 1.5mm,rounded corners, minimum width = 2cm, minimum height = 11cm, fill = darkgray, yshift = 1cm] (R_conv5){\rotatebox{90}{\HUGE 3$\times$3 conv, 4}};%
\node [right = of R_conv5.east,draw, line width = 1.5mm,rounded corners, minimum width = 2cm, minimum height = 11cm, fill = darkgray] (R_conv6){\rotatebox{90}{\HUGE 3$\times$3 conv, 4}};%
\node [right = of R_conv6.east,draw, line width = 1.5mm,rounded corners, minimum width = 2cm, minimum height = 11cm, fill = customgreen] (R_conv7){\rotatebox{90}{\HUGE 2$\times$2 conv, 3}};%

\draw [thick, line width = 1.5 mm] (R_unet.south west)+(0.2,0.5mm) -- (R_conv1.north west);
\draw [thick, line width = 1.5 mm] (R_unet.south east)+(-0.2,0.5mm) -- (R_conv7.north east);

\draw [-{Latex[length=8mm]}, thick, line width = 1.5mm] (R_conv1.east) -- (R_conv2.west);
\draw [-{Latex[length=8mm]}, thick, line width = 1.5mm] (R_conv2.east) -- (R_conv2.east-|R_mp.west);
\draw [-{Latex[length=8mm]}, thick, line width = 1.5mm] (R_mp.east) -- (R_conv3.west);
\draw [-{Latex[length=8mm]}, thick, line width = 1.5mm] (R_conv3.east) -- (R_conv4.west);
\draw [-{Latex[length=8mm]}, thick, line width = 1.5mm] (R_conv4.east) -- (R_up.west);
\draw [-{Latex[length=8mm]}, thick, line width = 1.5mm] (R_conv2.east-|R_up.east) -- (R_conv5.west);
\draw [-{Latex[length=8mm]}, thick, line width = 1.5mm] (R_conv5.east) -- (R_conv6.west);
\draw [-{Latex[length=8mm]}, thick, line width = 1.5mm] (R_conv6.east) -- (R_conv7.west);
\draw [dashed,-{Latex[length=8mm]}, thick, line width = 1.5mm,transform canvas={yshift=-6mm}] (R_conv2.north east) -- (R_conv5.north west);

\end{tikzpicture}} }
	\vskip\baselineskip
	\subcaptionbox{The STN localization network} {
	\scalebox{0.3}{\definecolor{red}{rgb}{1,0,0}%
\definecolor{green}{rgb}{0,1,0}%
\definecolor{blue}{rgb}{0,0,1}%
\definecolor{yellow}{rgb}{1,1,0}%
\definecolor{orange}{rgb}{1,0.647,0}%
\definecolor{gold}{rgb}{1,0.843,0}%
\definecolor{purple}{rgb}{0.627,0.125,0.941}%
\definecolor{gray}{rgb}{0.745,0.745,0.745}%
\definecolor{brown}{rgb}{0.647,0.165,0.165}%
\definecolor{navy}{rgb}{0,0,0.502}%
\definecolor{pink}{rgb}{1,0.753,0.796}%
\definecolor{seagreen}{rgb}{0.18,0.545,0.341}%
\definecolor{turquoise}{rgb}{0.251,0.878,0.816}%
\definecolor{violet}{rgb}{0.933,0.51,0.933}%
\definecolor{darkblue}{rgb}{0,0,0.545}%
\definecolor{darkcyan}{rgb}{0,0.545,0.545}%
\definecolor{darkgray}{rgb}{0.663,0.663,0.663}%
\definecolor{darkgreen}{rgb}{0,0.392,0}%
\definecolor{darkmagenta}{rgb}{0.545,0,0.545}%
\definecolor{darkorange}{rgb}{1,0.549,0}%
\definecolor{darkred}{rgb}{0.545,0,0}%
\definecolor{lightblue}{rgb}{0.678,0.847,0.902}%
\definecolor{lightcyan}{rgb}{0.878,1,1}%
\definecolor{lightgray}{rgb}{0.827,0.827,0.827}%
\definecolor{verylightgray}{rgb}{0.927,0.927,0.927}%
\definecolor{lightgreen}{rgb}{0.565,0.933,0.565}%
\definecolor{customgreen}{rgb}{0.796,0.890,0.824}%
\definecolor{lightyellow}{rgb}{1,1,0.878}%
\definecolor{black}{rgb}{0,0,0}%
\definecolor{white}{rgb}{1,1,1}%
\newcommand{\HUGE}{\fontsize{35}{60}\selectfont}%
\newcommand{\HHUGE}{\fontsize{45}{60}\selectfont}%
\renewcommand{\arraystretch}{3.5}%
\begin{tikzpicture}[every text node part/.style={align=center}]
\node at (0,0) [draw, line width = 1.5mm,rounded corners, minimum width = 2cm, minimum height = 8cm, fill = darkgray] (R_conv1) {\rotatebox{90}{\HUGE 5$\times$5 conv, 4}};%
\node [right = of R_conv1.east,draw, line width = 1.5mm,rounded corners, minimum width = 2cm, minimum height = 8cm, fill = lightgray] (R_mp1){\rotatebox{90}{\HUGE 1/3 max pooling}};%
\node [right = of R_mp1.east,draw, line width = 1.5mm,rounded corners, minimum width = 2cm, minimum height = 8cm, fill = darkgray] (R_conv2){\rotatebox{90}{\HUGE 3$\times$3 conv, 4}};%
\node [right = of R_conv2.east,draw, line width = 1.5mm,rounded corners, minimum width = 2cm, minimum height = 8cm, fill = lightgray] (R_mp2){\rotatebox{90}{\HUGE 1/2 max pooling}};%
\node [right = of R_mp2.east,draw, line width = 1.5mm,rounded corners, minimum width = 2cm, minimum height = 8cm, fill = darkgray, xshift=1cm] (R_fc1){\rotatebox{90}{\HUGE FC 32}};%
\node [right = of R_fc1.east,draw, line width = 1.5mm,rounded corners, minimum width = 2cm, minimum height = 8cm, fill = customgreen] (R_fc2){\rotatebox{90}{\HUGE FC 6}};%

\draw [-{Latex[length=8mm]}, thick, line width = 1.5mm] (R_conv1.east) -- (R_mp1.west);
\draw [-{Latex[length=8mm]}, thick, line width = 1.5mm] (R_mp1.east) -- (R_conv2.west);
\draw [-{Latex[length=8mm]}, thick, line width = 1.5mm] (R_conv2.east) -- (R_mp2.west);
\draw [-{Latex[length=8mm]}, thick, line width = 1.5mm] (R_mp2.east) -- (R_fc1.west);
\draw [-{Latex[length=8mm]}, thick, line width = 1.5mm] (R_fc1.east) -- (R_fc2.west);
%\draw [-{Latex[length=8mm]}, thick, line width = 1.5mm] (R_conv2.east) -- (R_conv2.east-|R_mp.west);
%\draw [-{Latex[length=8mm]}, thick, line width = 1.5mm] (R_mp.east) -- (R_conv3.west);
%\draw [-{Latex[length=8mm]}, thick, line width = 1.5mm] (R_conv3.east) -- (R_conv4.west);
%\draw [-{Latex[length=8mm]}, thick, line width = 1.5mm] (R_conv4.east) -- (R_up.west);
%\draw [-{Latex[length=8mm]}, thick, line width = 1.5mm] (R_conv2.east-|R_up.east) -- (R_conv5.west);
%\draw [-{Latex[length=8mm]}, thick, line width = 1.5mm] (R_conv5.east) -- (R_conv6.west);
%\draw [-{Latex[length=8mm]}, thick, line width = 1.5mm] (R_conv6.east) -- (R_conv7.west);
%\draw [dashed,-{Latex[length=8mm]}, thick, line width = 1.5mm,transform canvas={yshift=-6mm}] (R_conv2.north east) -- (R_conv5.north west);

\end{tikzpicture}}
	}
	\caption{The learning architecture is built upon an STN part and a slimmer version of u-net. For ease of visualization, some abbreviations have been introduced above. For instance, the label ``3$\times$3 conv, 4" indicates a convolutional layer with four output channels and a 3$\times$3 kernel, while ``1/2 max pooling" refers to a maximum pooling layer, which subsamples the input to half of its original size. Finally, ``3$\times$3 upconv, 4" represents an upconvolutional layer, which doubles the input size, and ``FC 32" denotes a fully connected layer with 32 units. In (a), the dashed arrow indicates the concatenation of the high resolution content with the upsampled information. For all 3$\times$3 convolutional layers, unitary zero-padding and a stride of 1 were used to retain the input size, as opposed to the last layer, where no padding and a stride of 2 halve the input to obtain 20$\times$20 force distribution matrices. In (b), the \textit{localization network} of the STN is shown, which learns a 6D affine transformation. No padding and a stride of 1 were used for the convolutional layers. In both (a) and (b), batch normalization and rectified linear unit activations were used at all convolutional and fully connected layers, with the exception of the respective output layers (in green).}
	\label{fig:learning_architecture}
	\vspace{-5mm}
\end{figure}

In this paper, the neural network architecture is largely inspired by u-net \cite{unet}, a well-known architecture widely employed for image segmentation tasks. The original version in \cite{unet} exhibited a fully convolutional structure, with a contracting path to extract context from the image patches and a symmetric expanding path (via upsampling) to assign a label to each pixel. Additionally, high resolution information was fed to the upsampled layers to perform pixel-wise regression. Here, the blocks inspired by u-net are placed after a spatial transformer network (STN \cite{stn}), which learns an affine transformation of the input features conditioned to the input itself, with the purpose of aligning the optical flow with the contact force distribution. The architecture is depicted in Fig.~\ref{fig:learning_architecture}. Note that the STN block only transforms the input image using the learned affine transformation, retaining the initial input size. 

\begin{figure}
 \centering
	\subcaptionbox{Original OF features}{
	\includegraphics[width = 0.4\columnwidth]{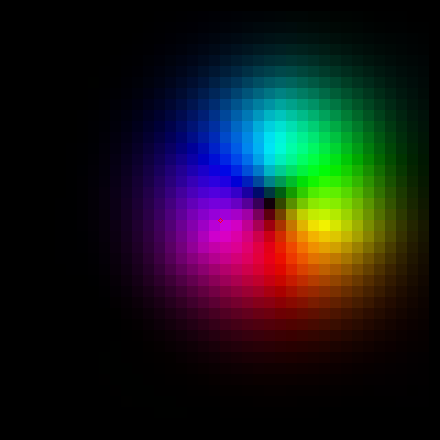}}
	\subcaptionbox{Deformed OF features}{
	\includegraphics[width = 0.4\columnwidth]{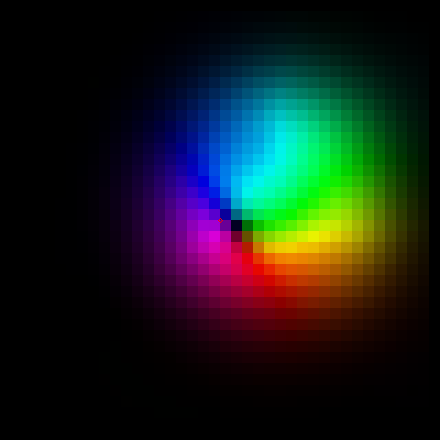}}\hfill
	%\scalebox{0.3}{\input{img/frames_definition.tex}}
	\caption{In this figure, an example of the synthetic optical flow (OF) features before (a) and after (b) elastic deformation is shown. The color represents direction, while darker regions represent smaller displacements.}
	\label{fig:elastic_deformation}
\end{figure}

In order to close the sim-to-real gap, the synthetic optical flow features are perturbed during training via elastic deformation noise (see Fig.~\ref{fig:elastic_deformation}), which has been proven to be especially suitable for pixel-wise regression tasks \cite{elastic_deformation}.

\section{REAL DATA ADJUSTMENT} \label{sec:real_data}
In real-world applications, the pinhole camera model does not capture the full camera projection, which exhibits lens distortion and various non-idealities. Camera calibration techniques aim to find the actual camera model from real images taken with the camera of interest. In this section, a procedure is proposed that enables the deployment of the neural network trained in simulation as described in \mbox{Section \ref{sec:sim_learning}} (assuming a pinhole projection) to a real-world sensor with a generic camera model. First, the employed camera calibration technique is presented. Then, an algorithm to remap the real world images to the pinhole reference model introduced in Section \ref{sec:sim_learning} is described. 

\subsection{Camera calibration}
\begin{figure}
	%\centering
	\subcaptionbox{Calibration procedure}{
		\includegraphics[width=.4\linewidth]{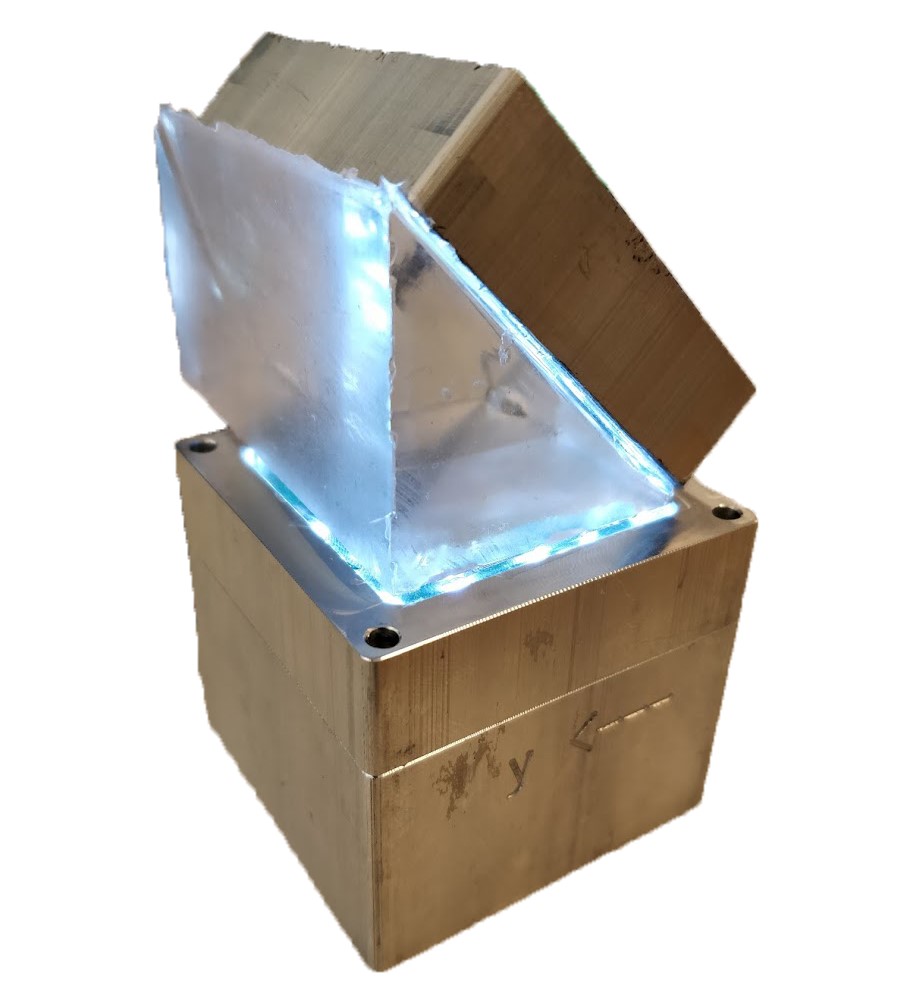} }
	\hfill
	\subcaptionbox{Calibration image} {
		\includegraphics[width=.55\linewidth]{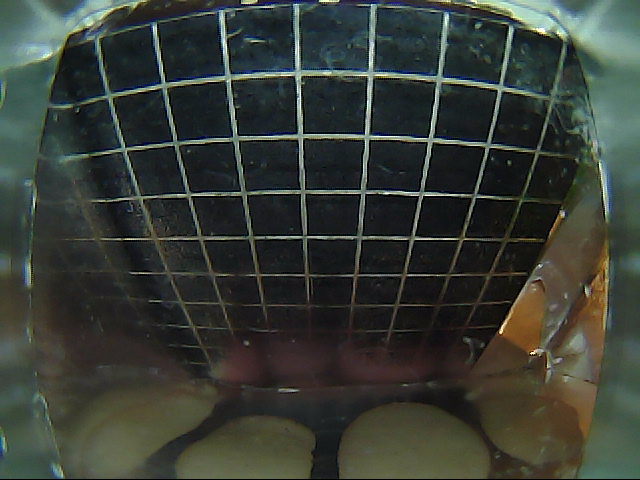}
	} 
	\caption{Six calibration images were used to obtain the real-world camera model. The images were taken before casting the two upper silicone layers, by placing a grid pattern at different positions, with only a transparent silicone medium of different shapes between the camera and the pattern. The calibration using the toolbox described in \cite{scaramuzza} obtained a subpixel reprojection error.}
	\label{fig:calibration}\vspace{-7mm}
\end{figure}

\begin{figure}
	%\centering
	%\includegraphics[width = 0.8\columnwidth]{example-image}
	\scalebox{0.3}{\tikzstyle{ipe stylesheet} = [
  ipe import,
  even odd rule,
  line join=round,
  line cap=butt,
  ipe pen normal/.style={line width=1},
ipe pen heavier/.style={line width=2},
ipe pen fat/.style={line width=3},
ipe pen ultrafat/.style={line width=6},
  ipe pen normal,
  ipe mark normal/.style={ipe mark scale=3},
ipe mark large/.style={ipe mark scale=13},
ipe mark small/.style={ipe mark scale=2},
ipe mark tiny/.style={ipe mark scale=1.1},
  ipe mark normal,
  /pgf/arrow keys/.cd,
  ipe arrow normal/.style={scale=7},
  ipe arrow large/.style={scale=10},
  ipe arrow small/.style={scale=5},
  ipe arrow tiny/.style={scale=3},
  ipe arrow normal,
  /tikz/.cd,
  ipe arrows, % update arrows
  <->/.tip = ipe normal,
  ipe dash normal/.style={dash pattern=},
  ipe dash dashed/.style={dash pattern=on 4bp off 4bp},
  ipe dash dotted/.style={dash pattern=on 1bp off 3bp},
  ipe dash dash dotted/.style={dash pattern=on 4bp off 2bp on 1bp off 2bp},
  ipe dash dash dot dotted/.style={dash pattern=on 4bp off 2bp on 1bp off 2bp on 1bp off 2bp},
  ipe dash normal,
  ipe node/.append style={font=\normalsize},
  ipe stretch normal/.style={ipe node stretch=1},
  ipe stretch normal,
  ipe opacity 10/.style={opacity=0.1},
  ipe opacity 30/.style={opacity=0.3},
  ipe opacity 50/.style={opacity=0.5},
  ipe opacity 75/.style={opacity=0.75},
  ipe opacity opaque/.style={opacity=1},
  ipe opacity opaque,
]
\definecolor{red}{rgb}{1,0,0}
\definecolor{green}{rgb}{0,1,0}
\definecolor{blue}{rgb}{0,0,1}
\definecolor{yellow}{rgb}{1,1,0}
\definecolor{orange}{rgb}{1,0.647,0}
\definecolor{gold}{rgb}{1,0.843,0}
\definecolor{purple}{rgb}{0.627,0.125,0.941}
\definecolor{gray}{rgb}{0.745,0.745,0.745}
\definecolor{brown}{rgb}{0.647,0.165,0.165}
\definecolor{navy}{rgb}{0,0,0.502}
\definecolor{pink}{rgb}{1,0.753,0.796}
\definecolor{seagreen}{rgb}{0.18,0.545,0.341}
\definecolor{turquoise}{rgb}{0.251,0.878,0.816}
\definecolor{violet}{rgb}{0.933,0.51,0.933}
\definecolor{darkblue}{rgb}{0,0,0.545}
\definecolor{darkcyan}{rgb}{0,0.545,0.545}
\definecolor{darkgray}{rgb}{0.663,0.663,0.663}
\definecolor{darkgreen}{rgb}{0,0.392,0}
\definecolor{darkmagenta}{rgb}{0.545,0,0.545}
\definecolor{darkorange}{rgb}{1,0.549,0}
\definecolor{darkred}{rgb}{0.545,0,0}
\definecolor{lightblue}{rgb}{0.678,0.847,0.902}
\definecolor{lightcyan}{rgb}{0.878,1,1}
\definecolor{lightgray}{rgb}{0.827,0.827,0.827}
\definecolor{lightgreen}{rgb}{0.565,0.933,0.565}
\definecolor{lightyellow}{rgb}{1,1,0.878}
\definecolor{black}{rgb}{0,0,0}
\definecolor{white}{rgb}{1,1,1}
\begin{tikzpicture}[ipe stylesheet]
  \pic[ipe mark large, green]
     at (272, 496) {ipe disk};
  \draw[ipe pen ultrafat]
    (192, 128)
     -- (192, 448);
  \draw[ipe pen ultrafat]
    (192, 448)
     -- (256, 448)
     -- (256, 448);
  \draw[ipe pen ultrafat]
    (256, 448)
     -- (256, 544)
     -- (704, 544)
     -- (704, 448);
  \draw[ipe pen ultrafat]
    (704, 448)
     -- (768, 448)
     -- (768, 448);
  \draw[ipe pen ultrafat]
    (768, 448)
     -- (768, 128);
  \draw[ipe dash dashed]
    (256, 448)
     -- (704, 448);
  \pic[ipe mark large, green]
     at (276.808, 524.2324) {ipe disk};
  \pic[ipe mark large, green]
     at (278.2292, 465.2516) {ipe disk};
  \pic[ipe mark large, green]
     at (362.7921, 464.541) {ipe disk};
  \pic[ipe mark large, green]
     at (318.7341, 463.8304) {ipe disk};
  \pic[ipe mark large, green]
     at (383.3998, 499.361) {ipe disk};
  \pic[ipe mark large, green]
     at (426.0365, 502.2034) {ipe disk};
  \pic[ipe mark large, green]
     at (445.9337, 459.5667) {ipe disk};
  \pic[ipe mark large, green]
     at (480.7537, 467.3834) {ipe disk};
  \pic[ipe mark large, green]
     at (509.1782, 465.9622) {ipe disk};
  \pic[ipe mark large, green]
     at (542.5769, 468.0941) {ipe disk};
  \pic[ipe mark large, green]
     at (477.9112, 497.2292) {ipe disk};
  \pic[ipe mark large, green]
     at (533.339, 526.3643) {ipe disk};
  \pic[ipe mark large, green]
     at (548.9724, 491.5443) {ipe disk};
  \pic[ipe mark large, green]
     at (577.3969, 520.6794) {ipe disk};
  \pic[ipe mark large, green]
     at (583.0818, 464.541) {ipe disk};
  \pic[ipe mark large, green]
     at (623.5867, 494.3867) {ipe disk};
  \pic[ipe mark large, green]
     at (651.3006, 498.6504) {ipe disk};
  \pic[ipe mark large, green]
     at (655.5643, 526.3643) {ipe disk};
  \pic[ipe mark large, green]
     at (689.6737, 525.6536) {ipe disk};
  \pic[ipe mark large, green]
     at (666.2235, 472.3577) {ipe disk};
  \pic[ipe mark large, green]
     at (687.5418, 459.5667) {ipe disk};
  \pic[ipe mark large, green]
     at (419.641, 531.3385) {ipe disk};
  \pic[ipe mark large, green]
     at (467.9627, 527.7855) {ipe disk};
  \pic[ipe mark large, green]
     at (411.1137, 465.2516) {ipe disk};
  \pic[ipe mark large, green]
     at (314.4704, 525.6536) {ipe disk};
  \pic[ipe mark large, green]
     at (308.0749, 490.123) {ipe disk};
  \pic[ipe mark large, green]
     at (384, 528) {ipe disk};
  \pic[ipe mark large]
     at (480, 192) {ipe disk};
  \draw[ipe pen fat]
    (480, 192)
     -- (352, 288);
  \draw[ipe pen fat]
    (480, 192)
     -- (608, 288);
  \draw
    (372.9707, 272.2719)
     -- (586.6296, 271.9722)
     -- (586.857, 272.1427);
  \draw[red, ipe pen ultrafat, ->]
(256, 448)
-- node [left,pos=1] { \scalebox{3.5}{$z^G$}} (256, 496);
\draw[red, ipe pen ultrafat, ->]
(256, 448)
-- node [below,pos=0.8]{ \scalebox{3.5}{$x^G$}} (304, 448);
\draw[red, ipe pen ultrafat, ->]
(480, 192)
-- node [left,pos=0.8]{ \scalebox{3.5}{$z^P$}} (480, 144);
\draw[red, ipe pen ultrafat, ->]
(480, 192)
-- node [below,pos=1]{ \scalebox{3.5}{$x^P$}}(528, 192);
  \pic[ipe mark large, green]
     at (512, 512) {ipe disk};
  \pic[ipe mark large, green]
     at (592, 496) {ipe disk};
  \pic[ipe mark large, green]
     at (608, 528) {ipe disk};
  \pic[ipe mark large, green]
     at (624, 464) {ipe disk};
  \pic[ipe mark large, green]
     at (352, 512) {ipe disk};
  \draw[ipe pen fat]
    (672, 160)
     -- (736, 272);
  \draw[ipe pen fat]
    (672, 160)
     -- (544, 224);
  \pic[ipe mark large]
     at (672, 160) {ipe disk};
  \draw
    (726.9943, 256.24)
     -- (561.9927, 215.0036);
  \draw[shift={(671.891, 159.941)}, rotate=18.4441, red, ipe pen ultrafat, ->]
    (0, 0)
     -- node [left,pos=1]{ \scalebox{3.5}{$z^C$}}(0, -48);
  \draw[shift={(671.891, 159.941)}, rotate=18.4441, red, ipe pen ultrafat, ->]
    (0, 0)
     -- node [right,pos=1]{ \scalebox{3.5}{$x^C$}} (48, 0);
  \draw[ipe dash dotted]
    (672, 160)
     -- (640, 448);
  \draw[ipe dash dotted]
    (640, 448)
     -- (480, 192);
  \draw[->]
    (480, 192)
     -- node [left,pos=0.8]{ \scalebox{3.5}{$t_z^{GP}$}} (480, 448);
  \pic[ipe mark large]
     at (663.055, 240.5046) {ipe disk};
  \pic[ipe mark large]
     at (530.1421, 272.2273) {ipe disk};
\end{tikzpicture}}
	\caption{The remapping procedure is shown in this figure. A new pinhole camera image is generated by filling each pixel with the corresponding real image pixel, found by reflecting the appropriate image ray on the bottom of the particle layer. Note that this approximation is exact when the origins of the pinhole (superscript $P$) and real (superscript $C$) camera coordinate systems coincide.}
	\label{fig:remapping}\vspace{-5mm}
\end{figure}
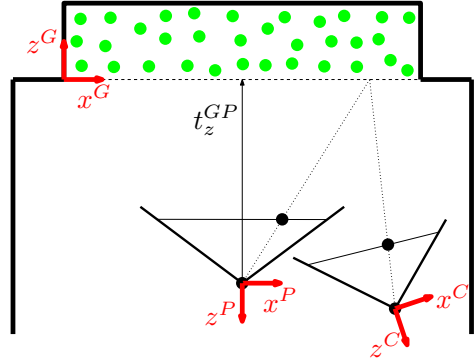

Given the large field of view of the fisheye lens employed, a calibration technique that accounts for the lens distortion is required to accurately match the camera projection. The strategy presented in \cite{scaramuzza} was employed in this paper, since it is tailored to omnidirectional and fisheye cameras and enables straightforward calibration via a \textsc{Matlab}\xspace toolbox. However, given the fact that in the application discussed here the camera is surrounded by silicone, the different refraction index with respect to air causes the light rays to deviate. Although the calibration method presented in \cite{scaramuzza} does not account for these refraction effects, shooting the calibration images directly through the same silicone medium (the stiffer rubber described in Section \ref{sec:sensing_principle}) yielded accurate results. One of the calibration images is shown in Fig.~\ref{fig:calibration}. As a result of the calibration, the toolbox provides a function denoted as \texttt{world2cam}, which accounts for the intrinsic parameters and projects a 3D point in the coordinate system of the real-world camera to the corresponding pixel in the image. By feeding an image in which the origin of the calibration pattern is aligned with the FEM coordinate system, the toolbox also outputs the extrinsic parameters of interest, $R^{GC}$ and $t^{GC}$. These parameters represent a transformation from the FEM (or gel) coordinate system to the coordinate system of the real-world camera (see Fig.~\ref{fig:remapping}).
% Figure with images as in Thomas report, in the caption mention how many figures and sub-pixel accuracy

\subsection{Images remapping}
The camera model obtained via calibration is employed to remap the images as if they were shot with the reference pinhole camera described in Section \ref{sec:sim_learning}. In this way, the neural network trained on the simulation data obtained via the pinhole model can be deployed to real tactile sensors with different intrinsic and extrinsic parameters, provided that they share the same gel geometry and mechanical properties. 

The procedure is based on reprojecting the pixels in the distorted image onto the pinhole image via the corresponding 3D world points. Since the images provide 2D information, only the direction of the corresponding 3D points can be retrieved. To overcome this limitation, the vertical coordinate of the 3D points is considered to be fixed and known. Since the majority of the particles visible in the image are the ones closer to the camera, this vertical coordinate is set to the lowest point $z^P := t_z^{GP}$ of the silicone layer containing the particles. Fig.~\ref{fig:remapping} depicts the remapping procedure. 

\begin{figure}
	\centering
	\subcaptionbox{Original image}{
	\includegraphics[height=.38\linewidth]{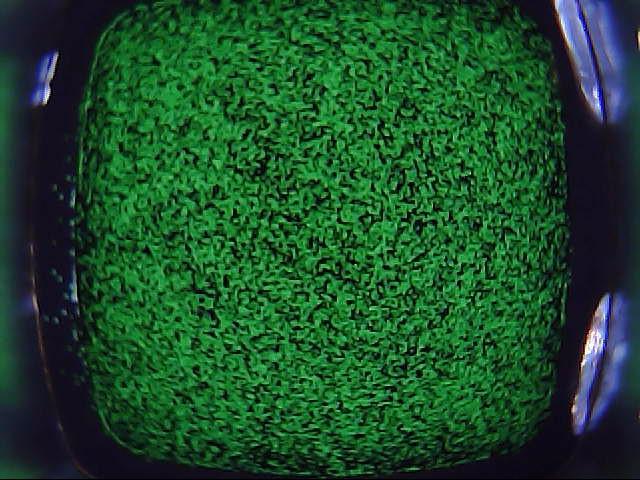} }
	\hfill
	\subcaptionbox{Remapped image} {
	\includegraphics[width=.38\linewidth]{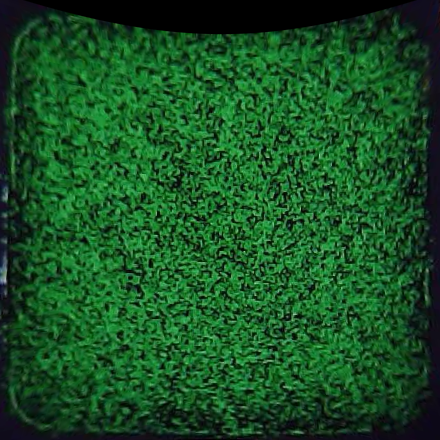}
	} 
	\caption{As shown in this figure, the remapping procedure removes most of the image distortion (see (a)), with the particle layer captured as a square in (b).}
	\label{fig:remapped_image}\vspace{-3mm}
\end{figure}

For each pixel $p := (u,v)$ in the pinhole image, the respective 3D world approximation $s^P := (x^P,y^P,t_z^{GP})$ is retrieved inverting the pinhole projection (see \eqref{eq:pos_projection2}) as
\begin{gather}
x^P = \dfrac{t_z^{GP}}{f}(u-u_\text{c}), \\
y^P = \dfrac{t_z^{GP}}{f}(v-v_\text{c}).
\end{gather}
The 3D point is then transformed to the coordinate system of the real-world camera via the appropriate rotation and translation,
\begin{align}
s^C = R^{GC}\left(R^{GP}\right)^{-1}\left(s^P - t^{GP}\right) + t^{GC}.
\end{align}
Note that the reference pinhole extrinsic parameters $R^{GP}$ and $t^{GP}$, which can be arbitrarily chosen, are set in the vicinity of the expected $R^{GC}$ and $t^{GC}$, respectively, to limit the impact of the approximation introduced above. These parameters depend on the design and assembly of the real-world tactile sensors. Finally, the corresponding pixel in the real-world image is retrieved via the \texttt{world2cam} function. An example of a remapped image is shown in Fig.~ \ref{fig:remapped_image}.

% Point to the figure, show the algorithm
% Mention that the approximation is reasonable for close origin
% Show image before and after
\section{RESULTS} \label{sec:results}
A fully synthetic dataset including 13,448 vertical indentations over the entire sensing surface was generated as described in Section \ref{sec:sim_learning}. A stainless steel, spherical-ended cylindrical indenter with a diameter of 10 mm was modeled and used for all the indentations. The simulations were of the same type as in \cite{sferrazza_fem}, where additional details are given.

The same setup was prepared in reality, where an equal indenter was attached to the spindle of an automatically controlled milling machine (Fehlmann PICOMAX 56 TOP). In this real-world scenario 200 indentations were performed, where the images were collected and matched to simulated ground truth labels, as described in \cite{sferrazza_fem}. These indentations were in the same range as the synthetic data and spanned a depth of 2 mm and normal forces up to 1.7 N. The simulation and real-world scenarios are depicted in Fig.~\ref{fig:flow_diagram}.

\begin{figure}
	%\centering
%	\includegraphics[width = 0.8\columnwidth]{example-image}
	\subcaptionbox{Multi-contact indentation}{
		\includegraphics[height=.3\linewidth]{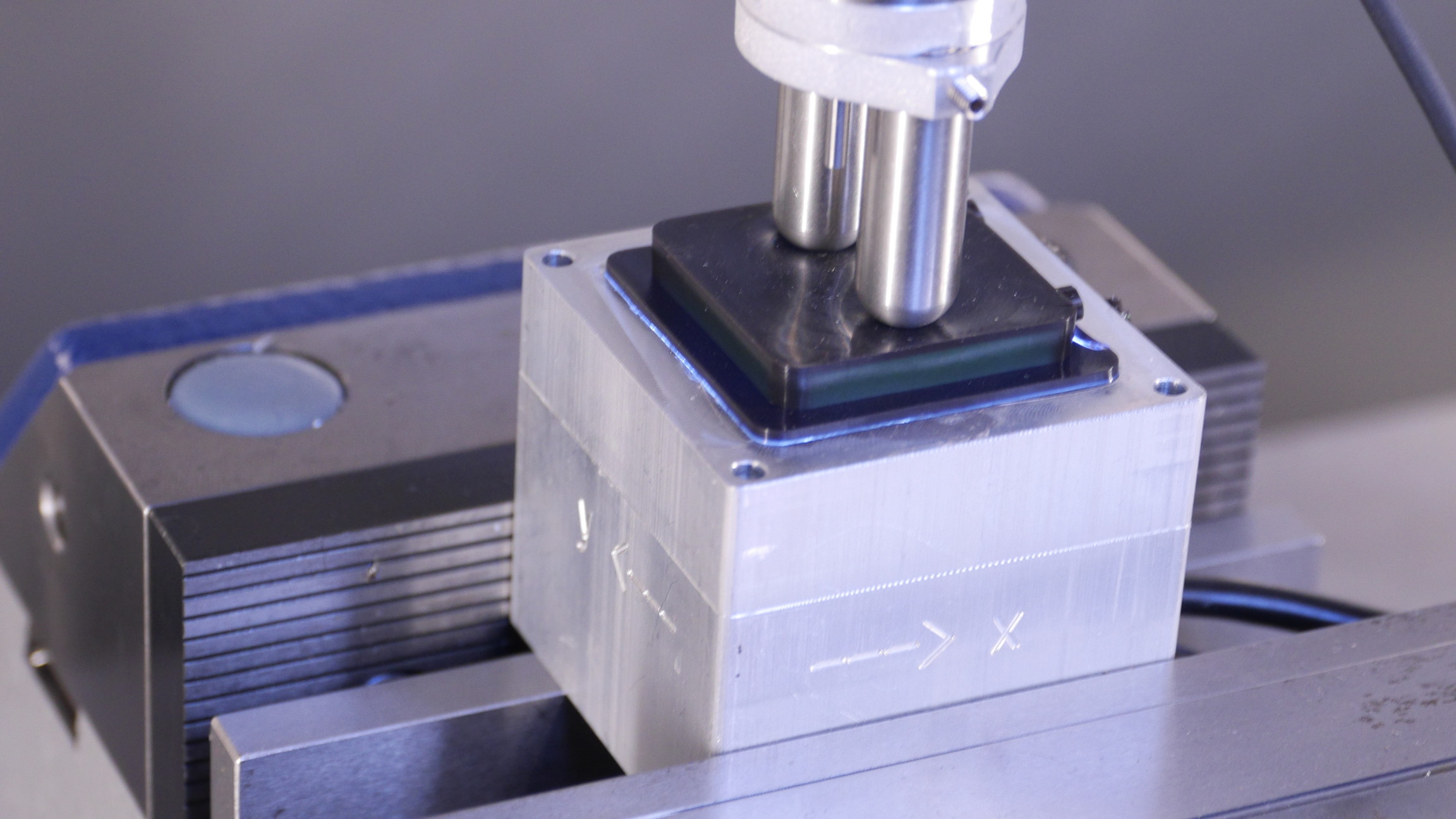} }
	\hfill
	\subcaptionbox{$F_z^G$ prediction} {
		\includegraphics[height=.3\linewidth]{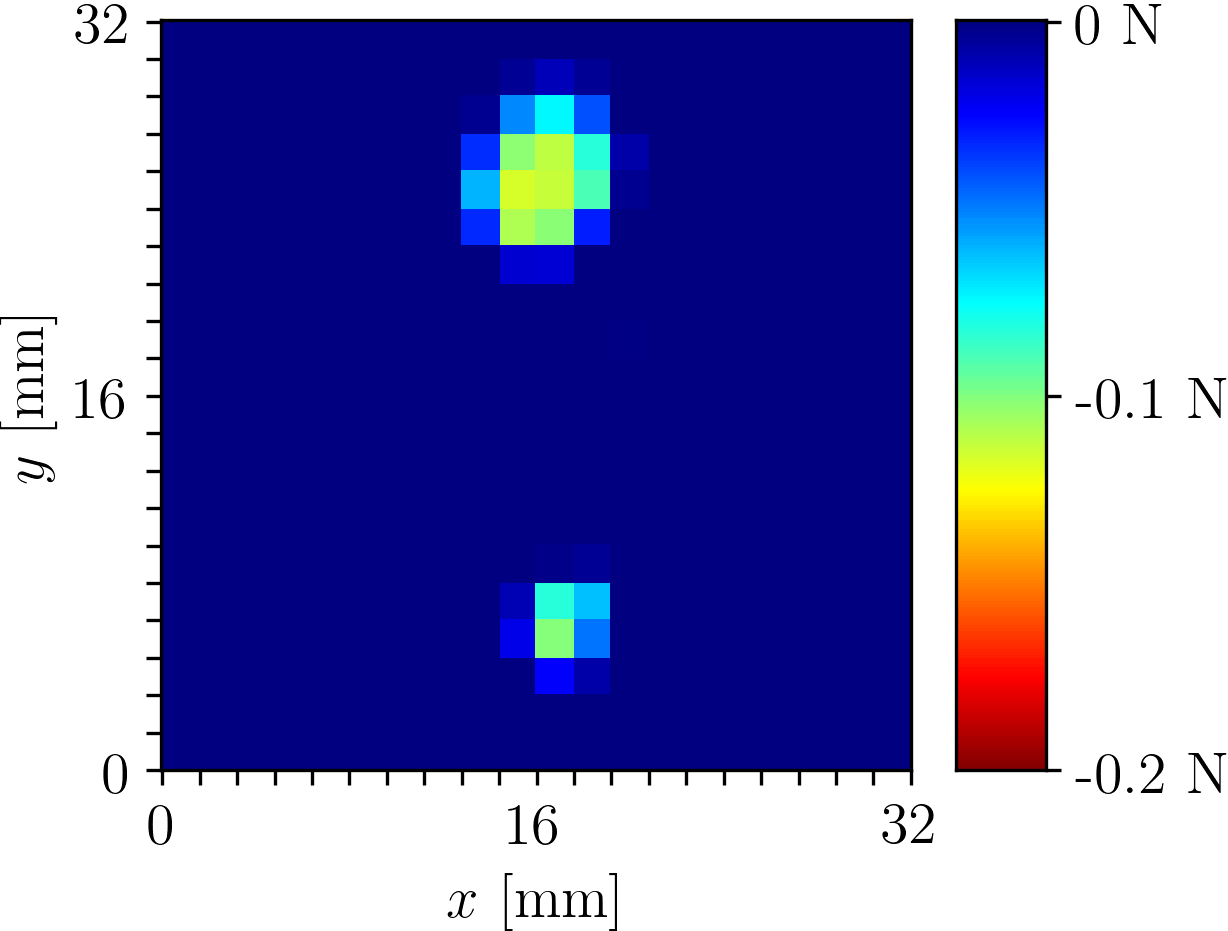}
	} 
	\caption{In (a), the indentation experiment involving multiple contact points is shown. The network fully trained on simulated single indentations detects both the distinct contact locations (see (b)), as well as the different pressure intensity (the two indenters have indeed a difference in length of 1 mm).}
	\label{fig:double_contact}	\vspace{-5mm}
\end{figure}

The neural network was trained on the fully synthetic dataset, by minimizing the mean squared error via the Adam optimizer \cite{adam_optimizer}, as implemented in Pytorch\footnote{https://pytorch.org/}. The network was then evaluated on the real-world data, composed of real images and respective synthetic labels. As described in Section \ref{sec:real_data}, the real-world images were remapped to the reference pinhole camera frame via the calibrated camera model, using both the intrinsic and extrinsic parameters. Additionally, a strategy was implemented to compensate for the mismatches introduced during production, assembly and calibration. In fact, in the current setup the exact alignment of the calibration pattern with the gel frame, which is necessary to retrieve the extrinsic parameters, is challenging. In order to take these deviations into account, a single real-world indentation taken in the center of the sensor at a depth of 1.25 mm was used to perform a local refinement of $t^{GC}$. This refinement was performed via a grid search around $t^{GC}$, by minimizing the mean squared error between the synthetic and real-world optical flow features. The magnitude of the resulting deviation was 0.7 mm, which is compatible with the hypothesis that this error was mainly introduced during the placement of the calibration pattern. A sample prediction on real-world data after refinement is shown in Fig.~\ref{fig:teaser}. The evaluation results on the full real-world dataset are shown in Table \ref{tab:results} for the cases with and without refinement. The RMSE rows show the root mean squared error on the respective components of the 3D contact force distribution. Additionally, the root mean squared error on the total force applied is shown, denoted as RMSET. The performance after the one-point refinement is comparable to the resolution of commercial force sensors, as was the case in \cite{sferrazza_fem} using real-world training images. The remaining gap is in large part due to artifacts introduced by the DIS algorithm (see Fig.~\ref{fig:teaser}) and to the modeling approximations. The convolutional nature of the learning architecture presented here exhibits promising generalization capabilities. In fact, the network was only trained on vertical indentations with a single, spherical-ended indenter, but exhibits sensible predictions for contacts with multiple bodies, as shown in Fig.~\ref{fig:double_contact}. The real-time prediction at 50 Hz on a standard laptop CPU is shown in the video attached to this paper, which also includes the prediction of the horizontal force distribution and the contact with objects of different shape.

\begin{table}[h!]
	\centering
	\begin{tabular}{c|c|c|c} 
		\hline
		\rule{0pt}{2.5ex}
		Metric & $F_x^G$ & $F_y^G$ & $F_z^G$\\
		\hline
		\rule{0pt}{3ex}
		RMSE & 0.002 N & 0.002 N & 0.005 N\\ 
		$\text{RMSET}$ & 0.032 N& 0.043 N& 0.150 N \\
		\hline
		\rule{0pt}{2.5ex}
		$\text{RMSE (refined)}$ & 0.001 N & 0.001 N & 0.004 N\\
		$\text{RMSET (refined)}$ & 0.032 N & 0.041 N & 0.131 N \\
		\hline
	\end{tabular}
\caption{Resulting errors on force distribution and total force\label{tab:results}	}
\end{table}
\vspace{-3mm}
\section{CONCLUSION} \label{sec:conclusion}
In this paper, a strategy has been presented to train an artificial neural network, which aims to reconstruct the three-dimensional contact force distribution applied to the soft surface of a vision-based tactile sensor. The generation of a fully synthetic dataset enables the training of the network in simulation, exhibiting accurate sim-to-real transfer when evaluated on real-world data. Additionally, the convolutional structure of the network facilitates generalization to a variety of contact conditions.

The remaining errors can mainly be explained by two factors: the discrepancies between real and synthetic optical flow features; and the fact that the elastic deformation noise injected during training and found to be essential for the sim-to-real transfer may excessively deteriorate the information relating to shear forces, which are rather small in the vertical indentation setup. Future work will focus on both these issues, by investigating appropriate noise characteristics that aim to emulate the imperfections of the real optical flow, and by augmenting the dataset with contact conditions that exhibit higher shear forces.


%%%%%%%%%%%%%%%%%%%%%%%%%%%%%%%%%%%%%%%%%%%%%%%%%%%%%%%%%%%%%%%%%%%%%%%%%%%%%%%%

%%%%%%%%%%%%%%%%%%%%%%%%%%%%%%%%%%%%%%%%%%%%%%%%%%%%%%%%%%%%%%%%%%%%%%%%%%%%%%%%

%%%%%%%%%%%%%%%%%%%%%%%%%%%%%%%%%%%%%%%%%%%%%%%%%%%%%%%%%%%%%%%%%%%%%%%%%%%%%%%%
\section*{APPENDIX}

\begin{figure}
	\hspace{-2.5cm}
	\scalebox{0.75}{\input{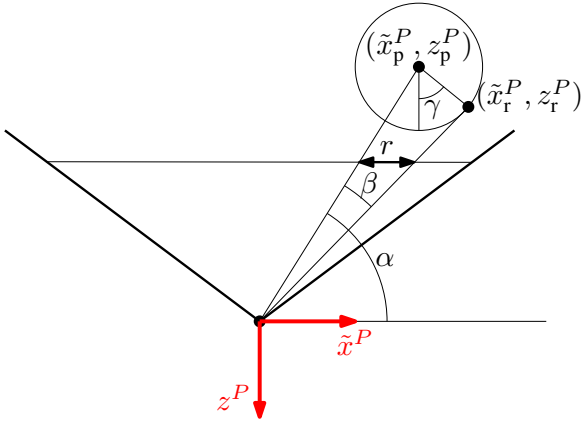}}
	\caption{As shown in this figure, the pixel radius $r$ on the image is computed by taking the difference between the pixel coordinate of the particle center and the projection of the point where the image ray is tangent to the particle.}
	\label{fig:appendix}
	\vspace{-5mm}
\end{figure}

Let $s_\text{p}^P := \left(x_\text{p}^P,y_\text{p}^P,z_\text{p}^P\right)$ be the position of a particle of radius $R$ in the pinhole camera frame. In general, a spherical particle is projected onto an ellipse in the image plane, which has its major axis in the plane containing the camera optical axis and the camera ray passing through the center of the sphere. A proof of this fact can be found in \cite{perspective_projection}. Therefore, the half-length $r$ of the major axis can be computed by a 2D analysis on this plane, see Fig.~\ref{fig:appendix}, via the following steps:
\begin{gather}
\tilde{x}_\text{p}^P = \sqrt{\left(x_\text{p}^P\right)^2+\left(y_\text{p}^P\right)^2},\\
\alpha = \arctan\left(-\dfrac{z_\text{p}^P}{\tilde{x}_\text{p}^P}\right), \\
\beta = \arcsin\left(\dfrac{R}{\sqrt{\left(\tilde{x}_\text{p}^P\right)^2+\left(z_\text{p}^P\right)^2}}\right), \\
\gamma = \alpha - \beta, \\
\tilde{x}_\text{r}^P = \tilde{x}_\text{p}^P+R\sin\gamma, \qquad
z_\text{r}^P = z_\text{p}^P+R\cos\gamma ,\\\label{eq:radius}
r = \left|f\left(\dfrac{\tilde{x}_\text{r}^P}{z_\text{r}^P} - \dfrac{\tilde{x}_\text{p}^P}{z_\text{p}^P} \right)\right|.
\end{gather}
In Section \ref{sec:sim_learning}, considering the small size of the particles, the projected ellipse is approximated as a circle with radius $r$, computed as in \eqref{eq:radius}.

%\vspace{-10mm}

\section*{ACKNOWLEDGMENT}

The authors would like to thank Michael Egli for his support in the sensor manufacture.

%%%%%%%%%%%%%%%%%%%%%%%%%%%%%%%%%%%%%%%%%%%%%%%%%%%%%%%%%%%%%%%%%%%%%%%%%%%%%%%%

\bibliographystyle{IEEEtran}
\bibliography{references}

\end{document}